\documentclass{article}

\usepackage{microtype}
\usepackage{graphicx}
\usepackage{hyperref}
\usepackage{subfigure}
\usepackage{multirow}
\usepackage{booktabs}% for professional tables
\usepackage{enumitem}

\usepackage{hyperref}

\usepackage[accepted]{icml2023}

\usepackage{amsmath}
\usepackage{amssymb}
\usepackage{mathtools}
\usepackage{amsthm}

\usepackage[capitalize,noabbrev]{cleveref}

\theoremstyle{plain}
\newtheorem{theorem}{Theorem}[section]
\newtheorem{proposition}[theorem]{Proposition}
\newtheorem{lemma}[theorem]{Lemma}

\theoremstyle{definition}
\newtheorem{definition}[theorem]{Definition}

\theoremstyle{remark}
\newtheorem{remark}[theorem]{Remark}
\newtheorem{example}[theorem]{Example}

\usepackage{cleveref}
\usepackage{tikz}

\usepackage{fixmath} % for \mathbold
\usepackage{scalerel}
\newlength\bshft
\def\fakebold#1{\ThisStyle{\ooalign{$\SavedStyle#1$\cr%
  \kern-\bshft$\SavedStyle#1$\cr%
  \kern\bshft$\SavedStyle#1$}}}

\DeclareMathOperator{\norm}{\mathrm{norm}}
\DeclareMathOperator{\subhom}{\mathrm{subhom}}
\DeclareMathOperator{\stronglysubhom}{\mathrm{s-subhom}}

\newcommand{\minval}{\alpha_1}
\DeclareMathOperator{\dom}{dom}
\newcommand{\maxval}{\alpha_2}

\DeclareMathOperator{\Diag}{Diag}
\def\one{\mathbf 1}
\let\plainTheta\Theta
\renewcommand{\Theta}{\mathit \plainTheta}

\icmltitlerunning{Subhomogeneous Deep Equilibrium Models}

\usepackage{xcolor}
\definecolor{shadecolor}{rgb}{0.85,0.85,0.85}

\begin{document}

\twocolumn[
\icmltitle{Subhomogeneous Deep Equilibrium Models}



\icmlsetsymbol{equal}{*}

\begin{icmlauthorlist}
\icmlauthor{Pietro Sittoni}{yyy}
\icmlauthor{Francesco  Tudisco}{yyy,edi}
\end{icmlauthorlist}

\icmlaffiliation{yyy}{School of Mathematics, Gran Sasso Science Institute, L'Aquila, Italy}
\icmlaffiliation{edi}{School of Mathematics and Maxwell Institute, University of Edinburgh, Edinburgh, UK}

\icmlcorrespondingauthor{Pietro Sittoni}{pietro.sittoni@gssi.it}
\icmlcorrespondingauthor{Francesco Tudisco}{f.tudisco@ed.ac.uk}

\icmlkeywords{Deep Learning, Implicit Neural Networks, DEQ, Fixed Points, Lipshitz Networks}

\vskip 0.3in
]



\printAffiliationsAndNotice{}  % leave blank if no need to mention equal contribution

\begin{abstract}
Implicit-depth neural networks have grown as powerful alternatives to traditional networks in various applications in recent years. However, these models often lack guarantees of existence and uniqueness, raising stability, performance, and reproducibility issues. In this paper, we present a new analysis of the existence and uniqueness of fixed points for implicit-depth neural networks based on the concept of subhomogeneous operators and the nonlinear Perron-Frobenius theory. Compared to previous similar analyses, our theory allows for weaker assumptions on the parameter matrices, thus yielding a more flexible framework for well-defined implicit networks. We illustrate the performance of the resulting subhomogeneous networks on feedforward, convolutional, and graph neural network examples.      
\end{abstract}

\section{Introduction}
% Example citation \cite{winston2020monotone}

Implicit-depth Neural Networks (NNs) have emerged as powerful tools in deep learning. Rather than using a sequence of nested layers, these models define feature embeddings as the solution to specific nonlinear equations. Two popular examples of implicit-depth networks are  Neural ODEs \cite{Chen2019NeuralODE}, whose output is calculated by solving an ordinary differential equation, and Deep Equilibrium (DEQ) Models \cite{bai2019deq}, which define the output in terms of a fixed-point equation. 
%Tracing back to early works on Hopofield networks and recurrent backpropagation \cite{Almeida1990ALR,Pineda87,hopfield1982neural}, see also \cite{weinan2017proposal,celledoni2021structure},  
These approaches have a number of advantages as (a) they have been shown to match or even exceed the performance of traditional NNs on several tasks, including time series and sequence modeling \cite{bai2019deq,rusch2021coupled}, and  (b) memory-wise, they are more efficient than traditional NNs as backpropagation is done analytically and does not require storage of internal weights, 
allowing to handle deep NN architectures more efficiently.

% In the past few years, various works in deep learning has shown the strength of implicit-depth networks, this models, instead of calculating the output by applying several nonlinear transformation to the input, calculates the output by solving a defined equation. The two most famous example of implicit-depth networks are: the Neural ODE \cite{Chen2019NeuralODE}, whose output is calculated by solving an ordinary differential equation, and the Deep Equilibrium (DEQ) Model \cite{bai2019deq}, which, to calculate the output, a fixed-point equation is solved. The core idea of the DEQ approach dates back to some work in recurrent backpropagation \cite{Almeida1990ALR}\cite{Pineda87}, and recently has been  shown that the DEQ can achieve the performance of the state-of-the-art models in the sequence modeling domain \cite{bai2019deq}.

DEQ models can be viewed as infinite-depth feed-forward neural networks, with weight tying, i.e.\ where the same transformation is used on each layer \cite{dabre2019recurrent,dehghaniuniversal}.  Indeed, in this type of model, the evaluation of the network is executed by solving a fixed-point equation $z = f_\theta(z;x)$ which can be thought of as the limit for the number of layers $n\to\infty$ of an $n$-deep network $z^{(n)} = f_\theta(z^{(n-1)};x)$. As the network is now implicitly defined as the solution of the equation $f_\theta(z;x)-z=0$, one can use the Implicit Function Theorem to compute and propagate the gradients, thus reducing the memory cost with respect to standard backpropagation \cite{bai2019deq,bai2020multiscale}.

% in implicit networks, the evaluation of the layer is executed by solving a fixed-point equation, then applying other layers to the fixed-point do not change its value. In addition, thanks to its implicit structure, the DEQ improve the memory efficiency, since the calculation of the gradient can be transformed into a fixed point equation thanks to the Implicit function Theorem, and by using a fixed-point solver, the memory consumption can be improved over the backpropagation algorithm.

Despite their potential advantages, not all DEQ models are well-defined. In fact, one of the main open questions for DEQ architectures is whether any fixed point actually exists and whether this is unique. Lack of uniqueness, in particular, can be a problem as for a given data $x$ and fixed pre-trained weights $\theta$, the resulting fixed-point embedding $z$ may change, raising stability, performance, and reproducibility issues. These potential drawbacks are often ignored or addressed only via empirical arguments based on experimental evidence that deep networks work well in practice. %: the   that very deep networks work is evidence that our training methods are already able to design layers that tend towards stable fixed points.
The most prominent line of analysis of uniqueness for deep equilibrium fixed points is based on monotone operator theory \cite{winston2020monotone,ryu2016primer}. While elegant and efficient, monotone operator DEQs (MonDEQs) require special parametrizations of the layer weights to guarantee the DEQ model $f_\theta$ is  operator monotone. 

In this work, we present a new analysis of existence and uniqueness of DEQ fixed points based on positive, subhomogeneous operators.
Using the Thomson projective metric and techniques from  nonlinear Perron--Frobenius theory \cite{lemmens_nussbaum_2012,gautier2019perron}, we provide a new theorem showing that a broad class of operators $f_\theta$ admits unique fixed points. 
In particular, we show that our uniqueness theorem holds for several example architectures of the form $f_\theta(z;x) = \sigma(Wz)+\mathrm{MLP}(x)$, provided the activation function is subhomogeneous, which we show is the case for a variety of commonly used activations. % with commonly used activation functions $\sigma$. %,  provided a suitable homogeneous layer normalization is added as final layer.  

%This result allows us to design stable DEQ models with a well-posed fixed-point equation which we illustrate how to practically implement.

This existence and uniqueness result allow us to design stable DEQ models under much weaker assumptions than available literature, avoiding any restriction on the learnable weight and using a large class of subhomogeneous activation functions.

%a SubDEQ and how to modify existing nets to ensure uniquenes. % We show some properties of the subhomegeneous operator suitable for application to  deep learning. Moreover, we show several examples of possible activation function for SubDEQ, which are commonly used in standard deep learning architectures. 
% We illustrates how to implement subhomegeneous operator to DEQ models. Since, in order to ensure convergence a normalization  step is required.

Our theoretical findings are complemented by several experimental evaluations where we compare simple fully-connected and convolutional DEQ architectures based on monotone operators with the newly introduced subhomogeneous deep equilibrium model (SubDEQ) on benchmark image classification tasks.

\section{Related work}

%\new{Cite somewhere the JMLR paper of Mateusz?}

% \paragraph{Well-posed deep equilibrium models } 
The classical approach to layer design in deep learning is based on explicitly defined function compositions and the corresponding computational graph. In contrast, implicit-depth approaches do not explicitly define the computational graph and define the model implicitly as the solution of a specific equation. The computational graph needs then to be extracted from the implicit formulation. NeuralODE is a popular example where the model is defined as the solution of an ordinary differential equation and backpropagation can be implemented via the adjoint method \cite{Chen2019NeuralODE}. Another example is the DEQ, where the model is defined as the fixed point of a nonlinear function and backpropagation can be done using the implicit function theorem \cite{bai2019deq}. However, DEQ may not be well-posed, as fixed points of nonlinear mappings may not exist or may not be unique. Several works in the literature propose variations of DEQ architecture with criteria for certified well-posedness using different methods, including monDEQ \cite{winston2020monotone} based on monotone operator theory, applying contraction theory \cite{jafarpour2021robust},  using an over-parametrized DEQ with a condition on the initial equilibrium point, or exploiting linear and nonlinear Perron-Frobenius theory on graph neural networks and autoencoders, \cite{gu2020implicit,ghaoui2020implicit,piotrowski2024fixed}.
%
% \paragraph{Robustness of deep equilibrium models} 
Just like standard explicit networks, DEQ models are vulnerable to adversarial input perturbations \cite{NEURIPS2022_43da8cca} and well-posed parameterizations may improve the robustness of the model. For example, using semialgebraic representation of monDEQ, \citet{chen2021semialgebraic} showed that monDEQ are more robust to $l^2$ perturbation. Similarly, \citet{wei2022certified} showed that unlike generic DEQs, monDEQ can achieve comparable $l^{\infty}$ certified robustness to similarly-sized fully explicit networks, via interval bound propagation. 
%
%It is well-known that neural networks can be vulnerable to adversarial input perturbations \cite{szegedy2014intriguing}, and also the DEQ are vulnerable, in fact \cite{NEURIPS2022_43da8cca} using white-box attacks elaborate defense strategies. Other works study in specific the robustness of monDEQ, using Semialgebraic Representation of Monotone Deep Equilibrium \cite{chen2021semialgebraic} suggest that monDEQ are more robust to $l^2$ perturbation respect to $l^{\infty}$ perturbation, on the other hand, \cite{wei2022certified} shows that monDEQ can achieve comparable $l^{\infty}$ certified robustness to similarly-sized fully explicit networks, via interval bound propagation.
%
% \paragraph{Deep equilibrium models in deep learning
% } 
Finally, we highlight that models based on DEQ architectures have been successfully employed in a wide range of specific applications. The first competitive performance was shown in the sequence modeling domain \cite{bai2019deq, bai2020multiscale}. Soon after that, efficient DEQ models have been designed for inverse problems in imaging \cite{gilton2021deep,zhao2023deep}, image denoising \cite{10070588}, optical flow estimation \cite{bai2022deep}, landmark detection \cite{micaelli2023recurrence}, semantic segmentation \cite{bai2020multiscale}, and generative modeling using a diffusion-based approach \cite{pokle2022deep}.
% \textcolor{purple}{
% Other applications of the nonlinear Perron-Frobenius theory to deep learning are for instance \cite{piotrowski2024fixed}, in which they use it to analyze nonnegative neural networks for understanding the behavior of autoencoders and propose future direction to study the wellposedness of the DEQ.
% }

\section{Subhomogeneous operators}\label{sec:subhom_operator}
The fundamental brick of our analysis is the notion of subhomogeneous operator. Recall that an operator $F:\mathbb R^n\to\mathbb R^n$ is (positively) $\mu$-homogeneous, for some $\mu>0$ if $F(\alpha z)= \alpha^\mu F(z)$ for all positive coefficients $\alpha>0$ and all entrywise positive vectors $z>0$. %Here and henceforth, all inequalities are meant componentwise. 
When $F$ is differentiable, Euler's theorem for homogeneous functions provides an elegant equivalent characterization of homogeneous operators as the set of $F$ such that the identity $F'(z)z=\mu F(z)$ holds for all $z$, where $F'(z)$ denotes the Jacobian of $F$ in $z$. The proposed notion of subhomogeneous operators generalizes the concept of homogeneous mappings, starting from this second characterization. %As a starting point of our work, we propose a set of operators, called subhomogeneus, defined as following.

\begin{definition}[Subhomogeneous operator]\label{def:sub}
Let $F\colon \mathbb{R}^n \to \mathbb{R}^m$ be a Lipschitz mapping. For a coefficient $\mu>0$ and $\Omega \subseteq \dom_+(F):=\{z \in \mathbb{R}^n \mid F(z)\geq0\}$, we say that $F$ is $\mu$-subhomogeneus in $\Omega$, briefly  $F\in\subhom_{\mu}(\Omega)$, if 
    \begin{equation}\label{def:inequality}
    |M z|  \leq \mu\,  F(z),
    \end{equation}
    for all $z \in \Omega$ and all $M \in \partial F(z)$, where $\partial F(z)$ denotes Clarke's generalized Jacobian of $F$ at the point $z$, and where all the inequalities, as well as the absolute value, are meant entrywise. Similarly, we say that $F$ is strongly $\mu$-subhomogeneus in $\Omega$, briefly  $F\in\stronglysubhom_{\mu}(\Omega)$, if 
      \begin{equation}\label{def:strongly-inequality}
    |M |\, |z|  \leq \mu\,  F(z),
    \end{equation}
    for all $z\in \Omega$ and all $M\in \partial F(z)$.
\end{definition}

Subhomogeneity generalizes both homogeneity and strong-subhomogeneity. In fact, all $\mu$-homogeneous differentiable operators are $\mu$-subhomogeneous due to Euler's theorem and, as $|Mz|\leq|M| |z|$  for all $z$, we immediately see that every strongly-subhomogeneus operator is subhomogeneus. However, homogeneity does not necessarily imply strong-subhomogeneity. This is shown for instance in Example \ref{ex:hom-not-stronglysubhom} below. 
On the other hand, we will see that strong subhomogeneity is preserved under composition while subhomogeneity is not, and this additional property will be useful to establish uniqueness results for brother families of deep equilibrium architectures.

\begin{example}\label{ex:hom-not-stronglysubhom}
Let $F\colon \mathbb{R}^2 \smallsetminus \{[0,0]\} \to \mathbb{R}^2$, be defined as $F(z)=F(x,y)=[\frac{x^2y}{x^2+y^2},0]$.  Clearly $F$ is $1$-homogeneus, with Jacobian given by
$$
F'(z)=
\begin{bmatrix}
\frac{2xy^3}{(x^2+y^2)^2} & \frac{x^2(x^2-y^2)}{(x^2+y^2)^2} \\
0 & 0
\end{bmatrix}\, .
$$
%and 
%$$
%|F'(z)|z = \begin{bmatrix}
%\frac{2x^2y^3}{(x^2+y^2)^2}+ \frac{x^2y(x^2-y^2)}{(x^2+y^2)^2} \\
%0 
%\end{bmatrix}.
%$$
Thus, $|F'(z)||z|$ calculated at $z=[1,2]$ is equal to $[0.56,0]$, while $F(z)=[0.25,0]$ in  $z=[1,2]$. This shows that $F$ is not strongly $1$-subhomogeneus in $\mathbb{R}^n_{++}$.
\end{example}

\begin{remark}\label{remark:sub_to_strongsub}
Note that subhomogeneity and strong subhomogeneity coincide when the mapping has a positive valued subgradient. In fact, if $F \in \subhom_{\mu}(\mathbb{R}^n_{+})$ and  $\partial F(z) \subseteq \mathbb R^n_+$ for all $z \in \mathbb{R}^n_{+}$, then $|M\,z|=M\,z=|M|\,|z|$ for all $z \in \Omega$ and all $M \in \partial F(z)$. Thus,  $F \in \stronglysubhom_{\mu}(\mathbb{R}^n_{+})$.
\end{remark}

%\begin{figure}    
%\centering
%\includegraphics[scale=0.35]{Preprint/icml_2024/subhom_diagram.pdf}
%    \caption{\ftnote{You may have fun doing this picture with tikz} %Relationship between homogeneous, subhomogeneous and strongly %subhomogeneous}
%    \label{fig:subhom_diagram}
%\end{figure}

% \begin{figure}[t]
%     \centering
%     \begin{tikzpicture}
%     \begin{scope} 
%     %\draw[draw = black] (-3.0,1.0) circle  (3.3);
%     %\draw[draw = black] (-2.6,1.6) circle (1.6);
%     %\draw[draw = black] (-3.6,0.25) circle (1.4);
%     %\draw[draw = black] (11.1,5.5) circle ++(0.3,0.3)
%     %\node at (-6.3,3.79) {$\boldsymbol{\mu}$\textbf{-subhomogeneus}};
%     \node at (-8.5,3.22) {$\boldsymbol{\mu}$\textbf{-subhomogeneus}};
%     \node at (-6.25,2.3){$\boldsymbol{\mu}$\textbf{-homogeneus}};
%     \node at (-9.2,-3.5) {\textbf{strongly}$\boldsymbol{\mu}$\textbf{-subhomogeneus}};
    
%     \draw[draw = black] (-12,-4) rectangle  (-4,3);
%     \draw[draw = black] (-11.25,-3.25) rectangle  (-7,0);
%     \draw[draw = black] (-8,-2.25) rectangle  (-5,2);
%     \end{scope}
%     \end{tikzpicture}
%     % \includegraphics{}
%     \caption{Relationship between homogeneous, subhomogeneous, and strongly subhomogeneous operators.\ftnote{Maybe remove this figure}}
%     \label{fig:enter-label}
% \end{figure}

In order to better understand the connection between homogeneity and the two proposed notions of subhomogeneity, we provide below an analogous of Euler's theorem for subhomogeneous operators, directly connecting the notion of subhomogeneous operator with the usual notion of homogeneity $F(\alpha z)= \alpha^\mu F(z)$, see also \cite{lemmens_nussbaum_2012}. The proof is deferred to  \Cref{proof_result}. 

\begin{proposition}\label{prop_eq_subhom_def}
Let $F\colon \mathbb{R}^n \to  \mathbb{R}^n$ be differentiable and Lipschitz. If $F \in \stronglysubhom_{\mu}(\mathbb{R}^n_{++})$. Then, $F \in \subhom_{\mu}(\mathbb{R}^n_{++})$ and for all  $\lambda \geq 1$ we have
$$
F(\lambda z) \leq \lambda^{\mu} F(z). 
$$
Assume moreover that $F'(z)>0$, for all $z>0$, i.e.\ the Jacobian $F'(z)$ is an entry-wise strictly positive matrix. 
Then, for  $\mu>0$, it holds
$$
F \in \subhom_{\mu}(\mathbb{R}^n_{++}) =\stronglysubhom_{\mu}(\mathbb{R}^n_{++})
$$
if and only if for all  $\lambda \geq 1$ we have
$$
F(\lambda z) \leq \lambda^{\mu} F(z).
$$
\end{proposition}

We provide below our main result showing uniqueness of fixed points for subhomogeneous operators, provided the homogeneity coefficient is small enough. 
Then, in Section \ref{sec:subhomdeq}, we will show that standard feed-forward linear layers of the form $\sigma(Wx)$ with a variety of broadly used activation functions $\sigma$ are indeed subhomogeneous, sometimes up to a minor modification.

\subsection{Main result: Existence and uniqueness of fixed points}
In this section, we show that $\mu$-subhomogeneus operators with small enough $\mu$ admit a unique fixed point.  %provided a suitable rescaling is implemented. 
All proofs are moved to \Cref{proof_result}. % via a 1-homogeneous functional.  

% Consider the cone of positive vectors  $\mathbb R_{++}^n=\{z\in\mathbb R^n:z>0\}$ and let $\varphi:\mathbb R^n\to\mathbb R$ be such that:
% \begin{itemize}[leftmargin=*,noitemsep,topsep=0pt]
%     \item (1-homogeneous) $\varphi(\alpha z)=\alpha\varphi(z)$ for all coefficients $\alpha>0$;
%     \item (positive)  $\varphi(z)>0$ for each $z>0$;
%     \item (order-preserving)   $\varphi(z)>\varphi(x)$ for each $z>x>0$.
% \end{itemize}

The proof of the existence and uniqueness of the fixed point is based on the Banach fixed-point theorem and the following fundamental completeness result % recalled in \Cref{thm:completeness} below.
\begin{theorem}[See e.g.\ \cite{lemmens_nussbaum_2012}]\label{thm:completeness}
    Let $\mathbb R_{++}^n:= \{z\in \mathbb R^n : z>0, \text{ entrywise}\}$ denote the interior of the nonnegative orthant. Consider the Thomson distance $\delta(x,y) = \|\ln(x)-\ln(y)\|_\infty$. Then, the pair $(\mathbb R_{++}^n, \delta)$ is a complete metric space.
\end{theorem}
Based on the theorem above, if we have an operator $F$ that maps $\mathbb R_{++}^n$ to itself and that is contractive with respect to $\delta$, then $F$ must have a unique fixed point in $\mathbb R_{++}^n$. 
Our main theorem below shows that this is always the case when $F$ is defined by a positive subhomogeneous operator.
We start by proving that if $F$ mapping  $\mathbb R_{++}^n$ to $\mathbb R_{++}^m$ is subhomogeneous, then it is Lispchitz continuous with respect to the Thompson distance, with a Lispchitz constant that depends only on its subhomogeneity degree.
\begin{theorem}\label{theorem:lip_unnorm}
Let $F: \mathbb{R}^n \to \mathbb{R}^m$ be a Lipschitz operator. Assume that $F$ is positive, i.e.\  $F(z)>0$ for all $z>0$, and that $F \in \subhom_{\mu}(\mathbb{R}_{++}^n)$ for some $\mu>0$. 
Then,
$$
\delta\left(F(x),F(y)\right) \leq \mu \,\delta(x,y),
$$
for all $x,y \in \mathbb{R}_{++}^n$.  
\end{theorem}
The following uniqueness result is now a direct consequence of  \Cref{thm:completeness} and \Cref{theorem:lip_unnorm}. 

\begin{theorem}\label{thm:fixed_point1}
Let  $F:\mathbb R^n\to\mathbb R^n$ be a Lipschitz operator.
Assume that $F$ is positive, i.e. $F (z) > 0$ for all $z > 0$, and be such that  $F\in \subhom_\mu(\mathbb R_{++}^n)$.
If $0<\mu<1$, then there exists a unique fixed point $F(z^*)=z^*\in \mathbb R_{++}^{n}$. Moreover, the sequence $z_{k+1}=F(z_k)$ converges to $z^*$ with linear convergence rate, namely 
\[
\|z_k-z^*\|_\infty \leq C \mu^k\, ,
\]
for any $z_0\in \mathbb{R}_{++}^{n }$.
\end{theorem}

In some occasions, for example when $F$ is matrix-valued as in the context of graph neural networks, one may need to normalize rows or columns of the hidden embedding to e.g.\ simulate a random walk. See also  \Cref{sec:graph_deq}. 

In order to show the uniqueness of fixed points for neural networks with this type of normalization layer,  we consider a general scaling function $\varphi:\mathbb R^n \to \mathbb R$ with the following properties:
\begin{itemize}[leftmargin=*,noitemsep,topsep=0pt]
    \item (1-homogeneous) $\varphi(\alpha z)=\alpha\varphi(z)$ for all coefficients $\alpha>0$;
    \item (positive)  $\varphi(z)>0$ for each $z>0$;
    \item (order-preserving)   $\varphi(z)>\varphi(x)$ for each $z>x>0$.
\end{itemize}

As $(\mathbb R^n_{++},\delta)$ is a complete metric space, then also $(\mathbb R^n_{++}/\varphi,\delta)$ must be complete, where $R_{++}^n/\varphi := \{z\in \mathbb R_{++}^n : \varphi(z)=1\}$. Our next result shows how \Cref{theorem:lip_unnorm} transfer to this setting
\begin{theorem}\label{theorem:lip}
Let $F: \mathbb{R}^n \to \mathbb{R}^m$ be as in \Cref{theorem:lip_unnorm}.  For a positive,  1-homogeneous, order-preserving functional $\varphi:\mathbb R^m\to\mathbb R$, define the $\varphi$-normalization layer map 
$$
\mathrm{norm}_\varphi:\mathbb{R}_{++}^n\to \mathbb{R}_{++}^m/\varphi, \qquad \mathrm{norm}_\varphi(z) = z/\varphi(z)
$$ 
and let 
$
G(z) = \mathrm{norm}_\varphi(F(z))\, .
$
Then,
$$
\delta\left(G(x),G(y)\right) \leq 2\mu \,\delta(x,y),
$$
for all $x,y \in \mathbb{R}_{++}^n$.  
Moreover, if $F$ is differentiable and its Jacobian matrix $F'(z)$ is entry-wise positive 
for all $z\in \mathbb R_{++}^n$, then 
$$
\delta\left(G(x),G(y)\right) \leq \mu \,\delta(x,y),
$$
for all $x,y \in \mathbb{R}_{++}^n$,
\end{theorem}
The following equivalent of \Cref{thm:fixed_point1} is now a relatively direct consequence. We formulate it explicitly for normalization layers acting column-wise, but we underline that its proof (see \Cref{proof_result}) can be easily adapted to different normalization patterns.
\begin{theorem}\label{cor:fixed_point2}
Let $G: \mathbb{R}^{n \times d} \to \mathbb{R}^{n \times d}$ be defined as
$$
G(x) = [\mathrm{norm}_{\varphi_1}(F_1(x)),\dots,\mathrm{norm}_{\varphi_d}(F_d(x))],
$$
where $F_i:\mathbb R^{n\times d}\to \mathbb R^{n}$ are $\mu$-subhomogeneous and $\varphi_i:\mathbb R^n \to \mathbb R$ are 1-homogeneous, positive, order-preserving functions. 
If $0<\mu<1/2$, then there exists a unique fixed point $G(z^*)=z^*$. Moreover, if all the $F_i$ are differentiable and their Jacobian matrices are entry-wise positive for all $z>0$, then a unique fixed point exists provided $0<\mu<1$. 
In both cases, the sequence $z_{k+1}=G(z_k)$ converges to $z$ with linear convergence rate, namely 
\[
\|z_k-z^*\|_\infty \leq C \mu^k\, ,
\]
for any $z_0\in \mathbb{R}_{++}^{n }$.
%
% Then, the same conclusion of \Cref{thm:fixed_point1} holds for $G$, provided the requirements for $F$ are satisfied by each~$F_i$. 
\end{theorem}

\section{Subhomogeneous deep equilibrium models}\label{sec:subhomdeq}
%\ftnote{Chiarire da subito qui che noi vogliamo dare la possibility di %considerare nonlinearita sia fuori che dentro. Ovviamente averle entrambe non ha %senso, ma potrebbe essere utile decidere di usare una piuttosto che l'altra e %quindi vogliamo studiare uniqueness properties per entrambe le scelte. Poi %prendendo $\sigma_i$ identita otteniamo ovvi corollari.}

Consider now the weight-tied, input-injected neural network 
\begin{equation}\label{eq:sequence}
    z^{(k+1)} = \sigma_1(\,\sigma_2(Wz^{(k)}) + f_\theta(x)\,)
\end{equation}
in which $x \in \mathbb R^d$
denotes the input, $z^{(k)}\in \mathbb R^n$ denotes the hidden unit at layer $k$, $\sigma_1, \sigma_2: \mathbb R\to \mathbb R$ denote  activation functions applied entrywise, $W\in \mathbb R^{n\times n}$ are the hidden unit weights, $f_\theta:\mathbb R^d\to\mathbb R^n$ is an input-injection embedding, e.g.\ defined by an MLP. We use here a fully connected layer for simplicity, but everything transfers unchanged to the convolutional case (i.e.\ if $Wz$ is replaced by $W\ast z$, with $W$ being the convolutional kernel). 
While in practice using both nonlinearities $\sigma_1$ and $\sigma_2$ may be redundant, we provide here a theoretical investigation of the model \eqref{eq:sequence} in its generality and we will then study specific architectures where either $\sigma_1=\mathrm{Id}$ or $\sigma_2=\mathrm{Id}$. %of the uniqueness p a theoretical perspective, we have the option to utilize both $\sigma_1$ and $\sigma_2$ as nonlinear activation functions, however for praticatal reason apply two activation function does not serve any meaningful purpose. Therefore, the architecture that we will propose later has either $\sigma_1=\mathrm Id$ and $\sigma_2$ as the non linear activation function or vice versa.}

  In the following, we show that a unique fixed point for \Cref{eq:sequence} exists, when the number of layers $k$ grows to infinity, provided the activation functions are subhomogeneous. 

Consider the following DEQ architecture
\begin{equation}\label{eq:SubDEQ}
    z = \sigma_1(\sigma_2(Wz)+f_\theta(x))
\end{equation}
or the corresponding normalized version
\begin{equation}\label{eq:SubDEQ_normalized}
    z = \norm_\varphi\big(\, \sigma_1(\sigma_2(Wz)+f_\theta(x))\, \big)
\end{equation}

where $\varphi$ is any positive, 1-homogeneous, order-preserving normalizing function (or multiple functions if the normalization layer is applied locally as in \Cref{cor:fixed_point2}).   
Note that $\varphi$ can be, for example, any standard $p$-norm. This type of layer normalization step is also used to e.g.\ reduce the network sensitivity to small perturbations  \cite{zhang2022rethinking, 
farnia2018generalizable} and could accelerate training \cite{ba2016layer,ioffe2015batch}.
%\ftnote{@piero, aggiungi quei lavori qui sopra}
% \textcolor{purple}{
% Alternatively, without the normalization layer \Cref{theorem:lip_unnorm} can be used to study the uniqueness of fixed point for
% \begin{equation}\label{eq:SubDEQ_unnormalized}
%     z =\sigma_1(\sigma_2(Wz)+f_\theta(x)))
% \end{equation}
% }

To study \eqref{eq:SubDEQ} and \eqref{eq:SubDEQ_normalized} using \Cref{thm:fixed_point1}, we now notice that it is enough to study the subhomogeneity of the activation functions $\sigma_1$ and $\sigma_2$. In fact, we can think of $F(z) := \sigma_1(\sigma_2(Wz)+f_\theta(x))$ as the composition of an activation function $\sigma_1$  applied entry-wise, with a translation $T(u)=u+f_\theta(x)$, another activation function $\sigma_2$, and a linear map $L(z) = Wz$ (or $L(z)=W\ast z$ for convolutional layers), 
\begin{equation}\label{eq:FF}
    F =  \sigma_1 \circ T \circ \sigma_2 \circ L \, . 
\end{equation}
Linear mappings are particular examples of homogeneous operators while translations are subhomogeneous. In the next lemma, we observe that the subhomogeneity of $F$ coincides with the subhomogeneity of $\sigma_1$ and $\sigma_2$, provided the input injection $f_\theta(x)$ is positive (entrywise). This requirement is not excessively restrictive, as it can be satisfied by using a positive nonlinear activation function, such as $\mathrm{ReLU}$, $\mathrm{SoftPlus}$, or by shifting bounded activations such as $\tanh$. %This positivity requirement on the input can be relaxed if we require strong subhomogeneity rather than just subhomogeneity. 
All proofs are moved to \cref{proof_result}. 
%\ftnote{Il motivo per cui usiamo sshom e' che la quarta %composizione fuori (con $\sigma_1$) richiedere sshom per poi darci %shom. La richiesta $y>0$ rimane.}

\begin{lemma}\label{lemma:subhom_composition}
Let $\Omega$ be a subset of  $\mathbb{R}^n$, $H$ be $h$-homogeneous, $P \in \subhom_{\mu}(H(\Omega))$, $T_y$ denote the translation by $y$, $T_y(z) =z+y$, and let $\boldsymbol Q \in \stronglysubhom_{\lambda}(T_y(P(H(\Omega)))).$
Then, if the following composition rules hold:
\begin{itemize}[leftmargin=*,noitemsep,topsep=0pt]
 \item $ P \circ H \in \subhom_{h \mu}(\Omega)$
 \item If $y\geq 0$, then $\boldsymbol Q \circ T_y \circ P \circ H \in \subhom_{h\mu\lambda }(\Omega)$
\end{itemize}
\end{lemma}

Thus, studying the subhomogeneity (resp.\ strong subhomogeneity) of $F$ in \eqref{eq:FF} boils down to the analysis of the subhomogeneity (resp.\ strong subhomogeneity) of $\sigma_1$ and $\sigma_2$. In particular, note that subhomogeneity is enough for $\sigma_2$ while strong homogeneity is required on $\sigma_1$ in order for the whole model to be subhomogeneous. However, we notice that in the typical case of activation functions acting entrywise this additional requirement is redundant. In fact, note that these subhomogeneity properties are inherited from the univariate function defining the activation layer when they act entrywise. Precisely, if $\sigma:\mathbb{R}\to \mathbb{R}$ is a Lipschitz function with $\sigma \in \subhom_{\mu}(\Omega)$, for $\Omega \subseteq \dom_{+}(\sigma)$, then we automatically have that the entrywise action of $\sigma$ on $\mathbb R^n$ is $\subhom_\mu(\Omega^n)$, with $\Omega^n=\Omega\times \cdots \times \Omega$, as the elements of the Clarke’s generalized Jacobian of $\sigma$ in $z\in \mathbb R^n$ are simply diagonal matrices whose $j$-th diagonal element is an element of the Clarke’s generalized Jacobian of $\sigma$ evaluated on the $j$-th component of~$z$. 
With the following remark, we can notice that the definitions of subhomogeneous and strongly subhomogeneous coincide in the univariate case. 
\begin{remark}\label{rem:univariate}
    Let $\sigma \colon \mathbb{R} \to \mathbb{R}$ be $\sigma \in \subhom_{\mu}(\Omega)$ for some $\mu>0$ and some $\Omega \subseteq \mathbb{R}$. Notice that, $|\sigma'(z)\,z|=|\sigma'(z)||z|$ for each $z \in \Omega$, this implies that $\sigma \in \stronglysubhom_{\mu}(\Omega).$
\end{remark}

In the next \Cref{sec:sub_act_fun} we will show a variety of examples of commonly used activation functions $\sigma$ that are subhomogeneous. For all such choices, the DEQ model \eqref{eq:SubDEQ} is well-defined as the existence of a unique equilibrium point $z$ is guaranteed. 

\section{Subhomogeneous activation functions with corresponding subhomogeneity coefficient}\label{sec:sub_act_fun}

\begin{table*}[t]
    \centering
    \resizebox{\textwidth}{!}{%
    \begin{tabular}{ccccccccc}
    \toprule
     Name    &  $\mathrm{Sigmoid}$ & $\mathrm{SoftPlus}$ & $\mathrm{Tanh}$ & $\mathrm{Tanh+1.2}$ & $\mathrm{Tanh+1.603}$ & $\mathrm{HardTanh}$ & $\mathrm{Leaky ReLU}$  & $\mathrm{Approxmax}$ \\
     \midrule
     $\mu$ & $1$ & $1$ & $1$ & $0.99$ & $0.499$ & $1$ & $1$ & $1$  \\
     $\Omega$ & $\mathbb{R}^n_+$ &$\mathbb{R}^n_+$  &$\mathbb{R}^n_{+}$ & $\mathbb{R}^n$ & $\mathbb{R}^n$ & $\mathbb{R}^n$ & $\mathbb{R}^n$  & $\mathbb{R}^n_+$   \\
    Differentiable & $\checkmark$ & $\checkmark$ & $\checkmark$ & $\checkmark$ & $\checkmark$ & & & $\checkmark$ \\
     \bottomrule
    \end{tabular}%
    }
    \caption{Examples of subhomogeneous activation functions}
    \label{tab:sub_act_fun}
\end{table*}

%\begin{tabular}{lccc}
%\hline Name & Offline & Ours & Bui et al. \\
%\hline ELEVATORS & $.50(.01)$ & $.57(.02)$ & $.57(.02)$ \\
%BIKE & $.24(.03)$ & $.44(.03)$ & $.49(.03)$ \\
%MAMMOGRAPHIC & $.40(.04)$ & $.41(.05)$ & $.42(.05)$ \\
%BANK & $.25(.03)$ & $.26(.03)$ & $.28(.04)$ \\
%MUSHROOM & $.00(.00)$ & $.02(.00)$ & $.08(.01)$ \\
%ADULT & $.32(.00)$ & $.35(.01)$ & $.36(.00)$ \\
%\hline
%\end{tabular}

We now propose several examples of subhomogeneous activation functions, they are well-known activation functions commonly used in deep learning.  All the proofs are moved to  \cref{proof_result}.

As a first example, we show that the $\mathrm{sigmoid}$ function is subhomogeneous, on the positive real axis, with $\mu=1$.
\begin{proposition}\label{lemma:sigmoid}
Let $\sigma:\mathbb{R} \to \mathbb{R}$ be defined as
$$
\sigma (z) = \mathrm{sigmoid}(z):= \frac{e^z}{1+e^z}.
$$
Then $\sigma \in \subhom_1(\mathbb{R_{+}})$.
\end{proposition}

Similarly to the $\mathrm{sigmoid}$, also the $\mathrm{SoftPlus}$ is subhomogeneous on 
$\mathbb{R_{+}}$ with $\mu=1$.

\begin{proposition}\label{lemma:softplus}
Let  $\sigma\colon \mathbb{R} \to \mathbb{R}$ be defined as
$$
\sigma(z) = \mathrm{softplus}(z):= \frac{1}{\beta}\ln(1+e^{\beta z}),
$$
where $\beta>0$. Then $\sigma \in \subhom_1(\mathbb{R_{+}})$.
\end{proposition}

%Other two examples of $1$-subhomogeneus operator are the $\mathrm{ReLU}$ and the $\mathrm{Leaky ReLU}$. While the $\mathrm{sigmoid}$ and the $\mathrm{SoftPlus}$ are $1$-subhomogeneus on $\mathbb{R}_+$, the $\mathrm{ReLU}$ and the $\mathrm{Leaky ReLU}$ are $1$-subhomogeneus on $\mathbb{R}_{++}$, this is due to the fact that they are strictly positive only on $\mathbb{R}_{++}$. As both activation functions are simply the identity on the positive orthant, they clearly are in $\subhom_1(\mathbb{R}_{++})$. 

Also, the hyperbolic tangent is subhomogeneous in $\mathbb{R}_+$.
%similar to the  $\mathrm{SoftPlus}$ and the $\mathrm{sigmoid}$, is $1$-subhomogeneus on the positive real axis.
\begin{proposition}\label{lemma:tanh}
Let  $\sigma \colon \mathbb{R}  \to \mathbb{R}$ defined as
$$
\sigma(z)=\tanh(z) := \frac{e^z-e^{-z}}{e^z+e^{-z}}.
$$
Then, $\sigma \in \subhom_1(\mathbb{R}_{+})$.
\end{proposition}

%\ftnote{Dovremmo chiarire quali di questi sono %anche \textbf{strongly} subhomogeneous}

The examples considered above are activations that are subhomogeneous only on the nonnegative/positive orthant. We provide below an example of subhomogeneous activation function that is subhomogeneous globally. This is obtained as a positive shift of the hyperbolic tangent. In fact, it is not difficult to notice that if $\sigma(z)=\tanh(z)+1+\epsilon$ with $\epsilon > 0$, then $\mu(\epsilon)= \max_z |z|\sigma'(z)\sigma(z)^{-1}$ is a decreasing function of $\epsilon$, i.e. $\mu(\epsilon_1)<\mu(\epsilon_2)$ for $\epsilon_1>\epsilon_2>0$. Moreover, it holds $\mu(\epsilon)<1$ for all   
$$
\epsilon > 0.199>\max_z \left\{|z|\mathrm{sech}^2(z)-\tanh(z)-1\right\} 
$$
and $\mu(\epsilon)<1/2$ for all 
$$
\epsilon > 0.602>\max_z \left\{2|z|\mathrm{sech}^2(z)-\tanh(z)-1\right\} \, .
$$
These computations directly lead to
\begin{proposition}\label{lemma:tanh+shift}
Let  $\sigma \colon \mathbb{R}\to\mathbb{R}$ be defined as
$$
\sigma(z) = \tanh(z)+\alpha\, .
$$
If $\alpha>1$, then $\sigma \in \subhom_\mu(\mathbb R)$. In particular, if $\alpha>1.2$ then $\mu<1$ and if $\alpha>1.602$ then $\mu<1/2$.  
\end{proposition}

Finally, similar to the hyperbolic tangent, we notice below that the non-differentiable piece-wise linear $\mathrm{hardtanh}$ activation function is $1$-subhomogeneus on $\mathbb{R}$. 
\begin{proposition}\label{lemma:hardtanh}
Let $\sigma:\mathbb{R} \colon \to \mathbb{R}$ be defined as
$$
\sigma(z) = \mathrm{hardtanh}(z):=
\begin{cases}
\minval  \quad \text{if} \quad z<\minval\\
\maxval  \quad \text{if} \quad z>\maxval \\
z  \quad \text{otherwise}
\end{cases}
$$
with $0<\minval<\maxval<\infty$. Then, $\sigma \in \subhom_1(\mathbb{R})$.
\end{proposition}

Other two examples of $1$-subhomogeneus operator over the whole real axis are the $\mathrm{ReLU}$ and the $\mathrm{Leaky ReLU}$ with a slope $\alpha \leq 0$, both are positive $1$-homogenous operator, thus they are also $1$-subhomogeneus in $\mathbb R$.

All activation functions considered above are univariate functions acting entrywise. Thus they are both subhomogeneous and strongly subhomogeneous as observed in \Cref{rem:univariate}. In the final example below we consider an example of a subhomogeneous function that is also strongly subhomogeneous but is not pointwise. %\ftnote{Pietro, per favore aggiungi la dimostrazione e controlla che valga anche $\sigma\in \stronglysubhom$. poi aggiungiamolo nello statement della proposizione}
\begin{proposition}\label{prop:softmax}
    Let $\sigma:\mathbb R^n \to \mathbb R$ be defined as 
    \[
    \sigma(z) = \mathrm{Approxmax}(z)=\ln \sum_i e^{z_i} \approx \max_i z_i
    \]
    Then, $\sigma \in \subhom_1(\mathbb R^n_+)$ and $\sigma \in \stronglysubhom_1(\mathbb R^n_+)$.
\end{proposition}

\subsection*{The power scaling trick} 
%\ftnote{Chiarire che questo trick e' utile se usiamo cose diverse dalla tanh shifted. Usare il lemma di composizione perche elevamento a potenza e' strongly subhom.}
\Cref{thm:fixed_point1} ensures a unique fixed point for any subhomogeneous operator $F$ exists, provided the subhomogeneity degree is small enough. As we have shown with the examples above, subhomogeneity is not a very stringent requirement, and many operators commonly used in deep learning are actually subhomogeneous without any modification. However, it is often the case that $\mu\geq 1$.
To reduce the degree of $\tanh$, we applied a positive shift $\alpha$. However, applying a scalar shift might not work all the time.   When the subhomogeneity constant is not small enough, a simple ``power scaling trick'' can be implemented for any subhomogeneous $F$. In fact, it follows directly from \Cref{lemma:subhom_composition} that if $F$ is $\mu$-subhomogeneous,  then $F^\alpha$ defined as the entrywise power $F^\alpha(x) = F(x)^\alpha$, is $\alpha\mu$-subhomogeneous, since the map $x \mapsto x^{\alpha}$ is strongly $\alpha$-subhomogeneous. Thus, to ensure uniqueness for \eqref{eq:SubDEQ} when $\sigma_i$ are $1$-subhomogeneous, %e.g.\ the sigmoid activation function, 
we can mildly perturb $\sigma_i$  into $\tilde \sigma_i(x) = \sigma_i(x)^{1-\varepsilon}$,  for any $\varepsilon>0$ arbitrary small. With this perturbed activation the DEQ in   \eqref{eq:SubDEQ} is guaranteed to have a unique fixed point.

\begin{figure}[t!]
    \centering
    \includegraphics[width=.49\columnwidth]{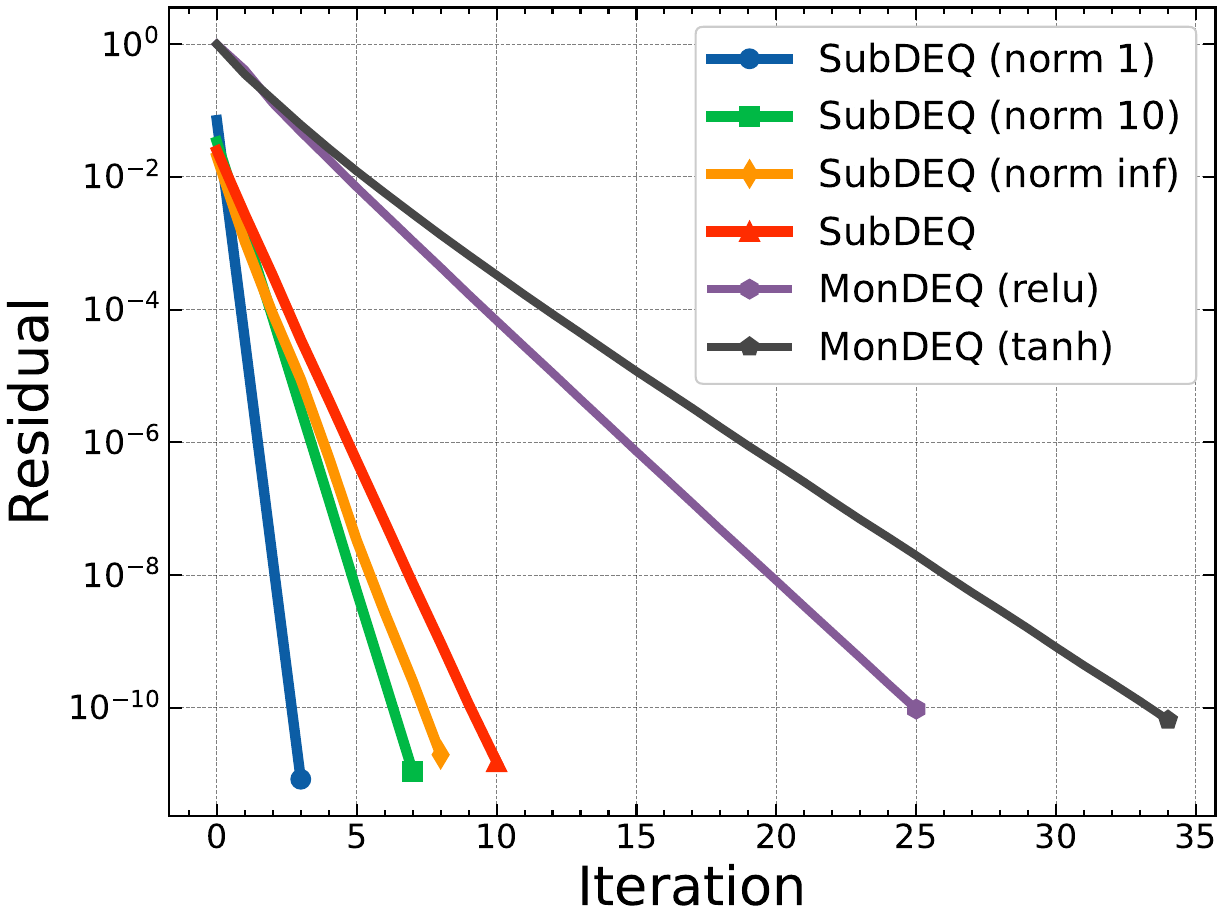} \includegraphics[width=.49\columnwidth]{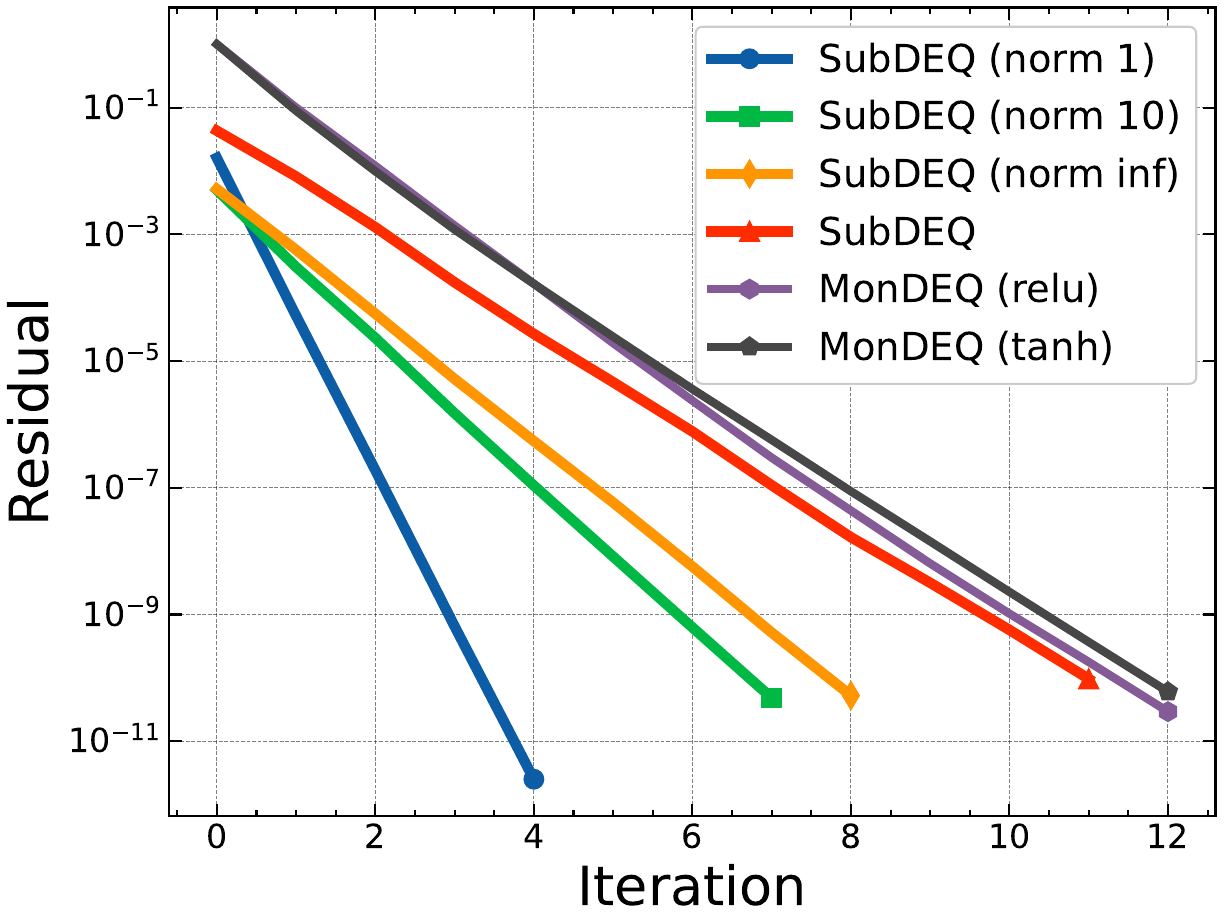}
    \caption{Iteration required by the fixed point method for SubDEQ vs  Peaceman-Rachford method for MonDEQ. Left: linear layer; Right: convolutional layer.}
    \label{fig:test_residueal_fc}
\end{figure}

\section{Experiments}\label{sec:experiments}
All the models are implemented in PyTorch and the code is available at \href{https://github.com/COMPiLELab/SubDEQ}{https://github.com/COMPiLELab/SubDEQ}.
We illustrate the performance of DEQ networks as in equations \eqref{eq:SubDEQ} and \eqref{eq:SubDEQ_normalized} on benchmark image datasets, as compared to alternative implicit neural networks, precisely Monotone operator-based DEQ (MonDEQ) \cite{winston2020monotone} and neural ordinary differential equations (nODE) \cite{Chen2019NeuralODE}, as well as standard explicit baseline architectures. 
%\ftnote{Far present che possiamo anche considerare $\sigma_2=Id$ e pero in quel caso ci %serve che $\sigma_1$ sia strongly subhomogeneous, quindi ad esempio $\tanh$ con matrice %$W$ positive e shift esterno $1.2$.}
As discussed in \Cref{sec:subhomdeq}, we will use only one $\sigma_i$ different from the identity function. To this end, we first consider the case $\sigma_1=\mathrm{Id}$, that is
$$
z = \sigma(Wz)+f_\theta(x)
$$
and 
$$
z = \mathrm{norm}_\varphi(\,\sigma(Wz)+f_\theta(x)\,).
$$
Thanks to \Cref{thm:fixed_point1}, \Cref{cor:fixed_point2}, and \Cref{lemma:subhom_composition} we can choose a suitable activation function $\sigma$ to guarantee the subhomogeneity of the architecture and thus the existence and uniqueness of the fixed point $z$, \emph{without any assumption on $W$}, we summarize in \ref{tab:summary of SubDEQ} all the possible combination of well-posed implicit-layers. Here we experiment with $\sigma(z) = \tanh(z)+\alpha$, with $\alpha$ chosen accordingly to \Cref{lemma:tanh+shift} to ensure a small enough subhomogeneity degree ($\mu<1$ for the standard model and $\mu<1/2$ for the normalized one) and thus uniqueness by \Cref{thm:fixed_point1}. Precisely, we 
    consider
    \begin{equation}\label{eq:subdeq_shiftedtanh}
    z = \tanh(Wz)+f_\theta(x)+1.2,
\end{equation}
and
\begin{equation}\label{eq:subdeq_shiftedtanh_normalized}
    z = \mathrm{norm}_{\|\cdot\|_{p}}(\,\tanh(Wz)+f_\theta(x)+1.603\,),
\end{equation}
where $1\leq p \leq + \infty$ and  $f_\theta(x)$ is a one-layer MLP with entry-wise positive final activation layer. As the architectures are now subhomogeneous and have a unique fixed point, we call \eqref{eq:subdeq_shiftedtanh} and \eqref{eq:subdeq_shiftedtanh_normalized} SubDEQ models.

%, and $1.603$ is meant as the vector with all the entries equal to  $1.603$. 
%Thanks to \Cref{lemma:subhom_composition}, we can substitute the weight matrices with kernels weight and use the convolution operator to have a convolutional implicit layer. \ftnote{Usiamo convoluzioni nell'architecture in pratica o no?}

% \usepackage{multirow}
\begin{table}[t]
\resizebox{1\columnwidth}{!}{
\begin{tabular}{ccccc}
\hline
\multirow{2}{*}{Model} & \multirow{2}{*}{$\norm_\varphi$} & \multicolumn{2}{l}{$\boldsymbol{\sigma} \in \subhom_{\boldsymbol{\mu}}(\boldsymbol{\Omega})$} & \multirow{1}{*}{Conditions} \\
\cline{3-4}
 & & $\boldsymbol{\mu}$ & $\boldsymbol{\Omega}$ &  on $\boldsymbol{W}$\\ 
\hline
\multirow{3}{*}{$\boldsymbol{\sigma}(\boldsymbol{W}\,z) + y$}                                              & \colorbox{shadecolor}{no}                & \colorbox{shadecolor}{$\boldsymbol{\mu}<1$} & \colorbox{shadecolor}{$\boldsymbol{\Omega} = \mathbb{R}^n$}  & \colorbox{shadecolor}{\textbf{None}}     \\
 & \multirow{2}{*}{yes}               & $\boldsymbol{\mu}<1/2$ & $\boldsymbol{\Omega} = \mathbb{R}^n$   & \textbf{None} \\
 &                & $\boldsymbol{\mu}<1$ & $\boldsymbol{\Omega} = \mathbb{R}^n_{++}$  & $\boldsymbol{W} \geq 0$ \\ \hline
\multirow{2}{*}{$\boldsymbol{\sigma}(\boldsymbol{W}\,z + y)$}                                              & no                & $\boldsymbol{\mu}<1$& $\boldsymbol{\Omega} = \mathbb{R}^n_{++}$ & $\boldsymbol{W} \geq 0$ \\
 & yes               & $\boldsymbol{\mu}<1$ & $\boldsymbol{\Omega} = \mathbb{R}^n_{++}$ & $\boldsymbol{W} \geq 0$ \\ \hline
\end{tabular}
% \begin{tabular}{clll}
% \hline
% \multicolumn{1}{l}{\begin{tabular}[c]{@{}l@{}}Model \\ ($y \geq 0$) \end{tabular}} & $\norm_\varphi$ & \begin{tabular}[c]{@{}l@{}}Constraints \\ on $\boldsymbol{W}$\end{tabular} & $\boldsymbol{\sigma} \in \subhom_{\boldsymbol{\mu}}(\boldsymbol{\Omega})$    \\ \hline
% \multirow{3}{*}{$\boldsymbol{\sigma}(\boldsymbol{W}\,z) + y$}                                              & \colorbox{shadecolor}{no}                & \colorbox{shadecolor}{\textbf{None}}                                                          & \colorbox{shadecolor}{$\boldsymbol{\mu}<1$; $\boldsymbol{\Omega} = \mathbb{R}^n$}      \\
%  & yes               & \textbf{None}                                                          & $\boldsymbol{\mu}<\frac{1}{2}$; $\boldsymbol{\Omega} = \mathbb{R}^n$     \\
%  & yes               & $\boldsymbol{W} \geq 0$                                                    & $\boldsymbol{\mu}<1$; $\boldsymbol{\Omega} = \mathbb{R}^n_{++}$  \\ \hline
% \multirow{2}{*}{$\boldsymbol{\sigma}(\boldsymbol{W}\,z + y)$}                                              & no                & $\boldsymbol{W} \geq 0$                                                    & $\boldsymbol{\mu}<1$; $\boldsymbol{\Omega} = \mathbb{R}^n_{++}$ \\
%  & yes               & $\boldsymbol{W} \geq 0$                                                    & $\boldsymbol{\mu}<1$; $\boldsymbol{\Omega} = \mathbb{R}^n_{++}$ \\ \hline
% \end{tabular}
}
\caption{Summary of requirements on weights and activations to guarantee existence and uniqueness of the DEQ fixed point, as well as the convergence of corresponding fixed point iteration. The setting requiring the fewest conditions is highlighted in gray color.}
\label{tab:summary of SubDEQ}
\end{table}

%\begin{table}[t]
%\begin{tabular}{lll}
%\hline
%Model                                           & $\mathbf{W}$                                %                                         & \boldmath$\sigma$                                  %                                                                   \\ \hline
%$\norm_{\varphi}(\boldsymbol{\sigma}(\mathbf{W}\,z) + f_{\theta}(x))$ & \begin{tabular}[c]%{@{}l@{}}$\mathbb{R}^n$\\ $\mathbb{R}^n_{+}$\end{tabular} & \begin{tabular}[c]%{@{}l@{}}$\subhom_{0.49}(\mathbb{R}^n)$\\ $\subhom_{0.99}(\mathbb{R}^n_{++})$\end{tabular} \\ %\hline
%$\norm_{\varphi}(\boldsymbol{\sigma}(\mathbf{W}\,z + f_{\theta}(x))$  & $\mathbb{R}^n_{+}$    %                                                      & $\subhom_{0.99}(\mathbb{R}^n_{++})$   %                                                                       \\ \hline
%$\boldsymbol{\sigma}(\mathbf{W}\,z) + f_{\theta}(x)$                  & $\mathbb{R}^n$        %                                                      & $\subhom_{0.99}(\mathbb{R}^n)$        %                                                                       \\ \hline
%$\boldsymbol{\sigma}(\mathbf{W}\,z + f_{\theta}(x))$                  & $\mathbb{R}^n_{+}$    %                                                      & $\subhom_{0.99}(\mathbb{R}^n_{++})$   %                                                                       \\ \hline
%\end{tabular}
%\caption{Summary of well-posed SubDEQ}
%\label{tab:summary of SubDEQ}
%\end{table}

\subsection{Efficiency of SubDEQ}

We now compare the convergence rate of the standard fixed--point method to find the equilibrium point of a SubDEQ, which is globally convergent due to \Cref{thm:fixed_point1}, with the convergence rate of the Peaceman-Rachford method to find the equilibrium point of a MonDEQ \cite{winston2020monotone}.
We analyze the implicit--layer of the SubDEQ in \eqref{eq:subdeq_shiftedtanh_normalized}
% $$
%     z = \mathrm{norm}_{\|\cdot\|_{p}}(\,(\tanh(Wz)+f_\theta(x))+1.603\,),
% $$
varying $p \in \{1,10,+ \infty\}$ and letting $f_\theta(x)=\mathrm{ReLU}(Ux+b)$ be a one-layer MLP with $\mathrm{ReLU}$ activation function.
We implement two different MonDEQs: the first one is defined as in \cite{winston2020monotone} via the following equation
\begin{equation}\label{eq:mondeq1}
    z = \mathrm{ReLU}(Wz+Ux+b)\, .
\end{equation}
The second one is defined as
\begin{equation}\label{eq:mondeq2}
    z = \tanh(Wz+ f_\theta(x))\,.
\end{equation}
to replicate the architecture of the proposed SubDEQ. 
%In order to guarantee that the MonDEQ architectures %have a unique fixed point, we restrict the weight %matrix $W$ in both the MonDEQ equations to satisfy %the operator monotone prametrization
%\[
%W=(1-m)I-A^\top A+B-B^\top
%\]
%with $A,B$ parameter matrices and $m=1$, c.f. %\cite{winston2020monotone}.

In order to guarantee that the MonDEQ architectures have a unique fixed point, we restrict the weight matrix $W$ in both the MonDEQ equations to satisfy the operator monotone parametrization
\[
W=A^\top A+B-B^\top
\]
with $A,B$ parameter matrices, c.f. \cite{winston2020monotone}.

For all models, as input, we choose $x \in \mathbb{R}^{128 \times 400}$  sampled from a uniform distribution on the interval $[0,1)$ using $\mathrm{torch.rand}$, the first dimension represents the batch--size and the second the dimension of the features space. The hidden embedding $z$ has a width of $150$.
%\ftnote{Quality of figures can/should be improved, let's talk about this in a meeting}
% \begin{figure}[t]
%     \centering
%     \includegraphics[width=\columnwidth]{figure_icml/res_test_fc.pdf}
%     \caption{Iteration required by fixed point method for dense SubDEQ and Peaceman-Rachford for dense MonDEQ.}
%     \label{fig:test_residueal_fc}
% \end{figure}
% \begin{figure}[h]
%     \centering
%     \includegraphics[width=\columnwidth]{figure_icml/res_test_conv.pdf}
%     \caption{Iteration required by fixed point method for convolutional SubDEQ and Peaceman-Rachford for convolutional MonDEQ.}
%     \label{fig:test_residueal_conv}
% \end{figure}

In \Cref{fig:test_residueal_fc} we plot the relative residual
$$
\|z_{k+1}-z_k\|_F/\|z_{k+1}\|_F,
$$
where $\|\cdot\|_F$ is the Frobenius norm and $z_k$ are the iterates of the fixed point methods. From \Cref{fig:test_residueal_fc}(left) we can notice that SubDEQ systematically requires fewer steps to converge with respect to MonDEQ. We also compare with the case of convolutional layers, instead of dense DEQ layers. In this setting, the input $x \in \mathbb{R}^{128 \times 1 \times 28 \times 28}$ is sampled from a uniform distribution on the interval $[0,1)$ using $\mathrm{torch.rand}$, the first dimension represent the batch--size, the second the number of the image channels and the last two the size of the image. The hidden fixed point embedding $z$ has $40$ channels and each channel has a size of $ 28 \times 28$. The relative residual of the iterations is shown in \Cref{fig:test_residueal_fc}(right). Also in this case, we can notice that SubDEQ systematically requires fewer steps to converge to converge than MonDEQ.

\subsection{SubDEQ on benchmark image datasets}\label{sec:experiment_image_dataset}

We now show the capacity of SubDEQ in terms of classification tasks. We train them on different image benchmark datasets: CIFAR-10 \cite{krizhevsky2009learning}, SVHN \cite{SVHN}, and MNIST \cite{lecun-mnisthandwrittendigit-2010}. We compare SubDEQ with  MonDEQ as well as neural ode (nODE) architectures \cite{Chen2019NeuralODE} and standard explicit network baselines. %\ftnote{Here and everywhere, Node should be changed in nODE} 
Overall, we consider the following models (for the feedforward implicit layer case):

\begin{enumerate}[leftmargin=*, noitemsep, topsep=0pt]
    \item SubDEQ (Normalized Tanh): $\mathrm{norm}_{\|\cdot\|_{\infty}}(\tanh(Wz)+ f_\theta(x) +1.603)$
    \item SubDEQ (Tanh):  $\tanh(Wz)+ f_\theta(x)+1.2$
    \item MonDEQ (ReLU): $z = \mathrm{ReLU}(Wz+Ux+b)$
    \item MonDEQ (Tanh): $z = \tanh(Wz+ f_\theta(x))$
    \item nODE (ReLU): $\dot{z}(t) = \mathrm{ReLU}(Wz(t))$, $z(0) = f_\theta(x)$
    \item nODE (Tanh): $ \dot{z}(t) = \tanh(Wz(t))$, $z(0) = f_\theta(x).$
\end{enumerate}

\begin{table}[t!]
\begin{tabular}{ll}
\hline
\textbf{Model}                                    & \textbf{Error \%}                        \\ \hline
\multicolumn{2}{l}{\textbf{MNIST (Dense)}}                                                                    \\ \hline
SubDEQ ($\mathrm{\text{Normalized Tanh}}$)                                & $2.088 \pm 0.1405$ \%  \\
\textbf{SubDEQ} ($\mathrm{\textbf{Tanh}}$)                         & $\mathbf{1.92 \pm 0.102}$  \textbf{\%}                                     \\
nODE ($\mathrm{ReLU}$)                                             & $2.356 \pm 0.0689$ \%                                     \\
nODE ($\mathrm{Tanh}$)                                             & $3.296 \pm  0.1082$ \%                                    \\
MonDEQ ($\mathrm{ReLU}$)                                           & $2.056 \pm 0.0484$ \%                                     \\
MonDEQ ($\mathrm{Tanh}$)                                           & $2.736 \pm 0.7491$ \%                                       \\
Standard MLP ($\mathrm{Tanh}$)                                     & $2.052 \pm 0.1452$ \%                                       \\
\hline
\multicolumn{2}{l}{ \textbf{MNIST (Convolutional)}}                                                            \\ \hline
SubDEQ ($\mathrm{\text{Normalized Tanh}}$)                                & $1.354 \pm 0.98$ \%                                      \\
SubDEQ ($\mathrm{Tanh}$)                                           & $0.706 \pm 0.011$ \%                                     \\
nODE ($\mathrm{ReLU}$)                                             & $1.184 \pm 0.3845 $ \%                                   \\
nODE ($\mathrm{Tanh}$)                                             & $0.826 \pm 0.0432 $ \%                                   \\
\textbf{MonDEQ} ($\mathrm{\textbf{ReLU}}$)                         & $\mathbf{0.654 \pm 0.0662}$  \textbf{\%}                   \\
MonDEQ ($\mathrm{Tanh}$)                                           & $1.096 \pm 0.0589 $ \%                                      \\
Standard CNN ($\mathrm{Tanh}$)                                     & $0.876 \pm 0.0739 $ \%                                      \\ 
\hline
\multicolumn{2}{l}{\textbf{CIFAR-10}}                                                                        \\ \hline
SubDEQ ($\mathrm{\text{Normalized Tanh}}$)                                & $28.364 \pm 0.377$ \%                                     \\
SubDEQ ($\mathrm{Tanh}$)                                           & $27.946 \pm 1.7564 $ \%                          \\
nODE ($\mathrm{ReLU}$)                                             & $33.58 \pm 1.2882 $ \%                                     \\
nODE ($\mathrm{Tanh}$)                                             & $28.792 \pm 1.3343$ \%                                     \\
\textbf{MonDEQ} ($\mathrm{\textbf{ReLU}}$)                         & $\mathbf{24.414 \pm 0.6521}$    \textbf{\%}                \\
MonDEQ ($\mathrm{Tanh}$)                                           & $35.618 \pm 1.1766$ \%                                     \\ 
Standard CNN ($\mathrm{Tanh}$)                                     & $27.157 \pm 0.4154 $ \%                                      \\ 
\hline
\multicolumn{2}{l}{\textbf{SVHN}}                                                                            \\ \hline
\textbf{SubDEQ} ($\mathrm{\textbf{Normalized Tanh}}$)              & $\mathbf{9.3562 \pm 0.2122}$                  \textbf{\%} \\
SubDEQ ($\mathrm{Tanh}$)                                           & $10.3987 \pm 0.41296 $ \%                                   \\
nODE ($\mathrm{ReLU}$)                                             & $33.8253 \pm 11.3008 $ \%                                  \\
nODE ($\mathrm{Tanh}$)                                             & $22.277 \pm 2.8740 $ \%                                  \\
MonDEQ ($\mathrm{ReLU}$)                                           & $11.0356 \pm 0.319 $ \%                                   \\
MonDEQ ($\mathrm{Tanh}$)                                           & $17.8849 \pm 0.9747 $ \%                                    \\ 
Standard CNN ($\mathrm{Tanh}$)                                     & $12.7243 \pm 0.1024 $ \%                                      \\ 
\hline
\multicolumn{2}{l}{\textbf{TinyImageNet}}                                                                            \\ \hline
\textbf{SubDEQ} ($\mathrm{\textbf{Normalized Tanh}}$)              & $\mathbf{70.1791 \pm 0.3666}$                  \textbf{\%} \\
SubDEQ ($\mathrm{Tanh}$)                                           & $74.6633 \pm 0.1511 $ \%                                   \\
MonDEQ ($\mathrm{ReLU}$)                                           & $85.49 \pm 0.2406 $ \%                                  \\
Standard CNN ($\mathrm{Tanh}$)                                     &  $73.76 \pm 1.4323 $ \%  \%                                  \\
\hline
\end{tabular}
 \caption{Mean $\pm$ std of the misclassification error on test set }\label{tab:reuslt_deq}
\end{table}

%\new{add details  about the explicit networks?}

We also consider a convolutional variant of these implicit layers, the only difference with the above dense layers is in the weights matrices, which we substitute with the standard convolutional kernels. For the standard explicit network baseline, we replace the implicit layer with a standard one using the same hyperparameter and $\tanh$ as the activation function, in both, the dense and the convolutional case.
For the normalized SubDEQ as in \eqref{eq:subdeq_shiftedtanh_normalized}, we decide to normalize each element of the batch for the feedforward, while with the convolutions we normalize along each row. For the nODE, we integrate the ODE over the interval $[0,1]$, as the output of the implicit layer we took the solution at time $t=1$.
%To have a fairer comparison among %the models, we normalize the %output of the implicit--layers of %the MonDEQ and the nODE using the %same norm as the SubDEQ. %\ftnote{For nODE we need to say at what $t$ we evaluate the network.}
For all the models,
we apply 1d batch normalization for the feedforward layers and 2d batch normalization for the convolutional layers. Moreover, as a last step, we feed the embedding into a softmax classifier, but with the convolutional architectures, before it, we apply an average pooling and then we flatten the tensor. We describe all the details about the hyperparameters in  \Cref{tab:model_hyper} in \Cref{app:additional-data-results}. 

\begin{figure}[t!]
    \centering    \includegraphics[width=1\columnwidth]{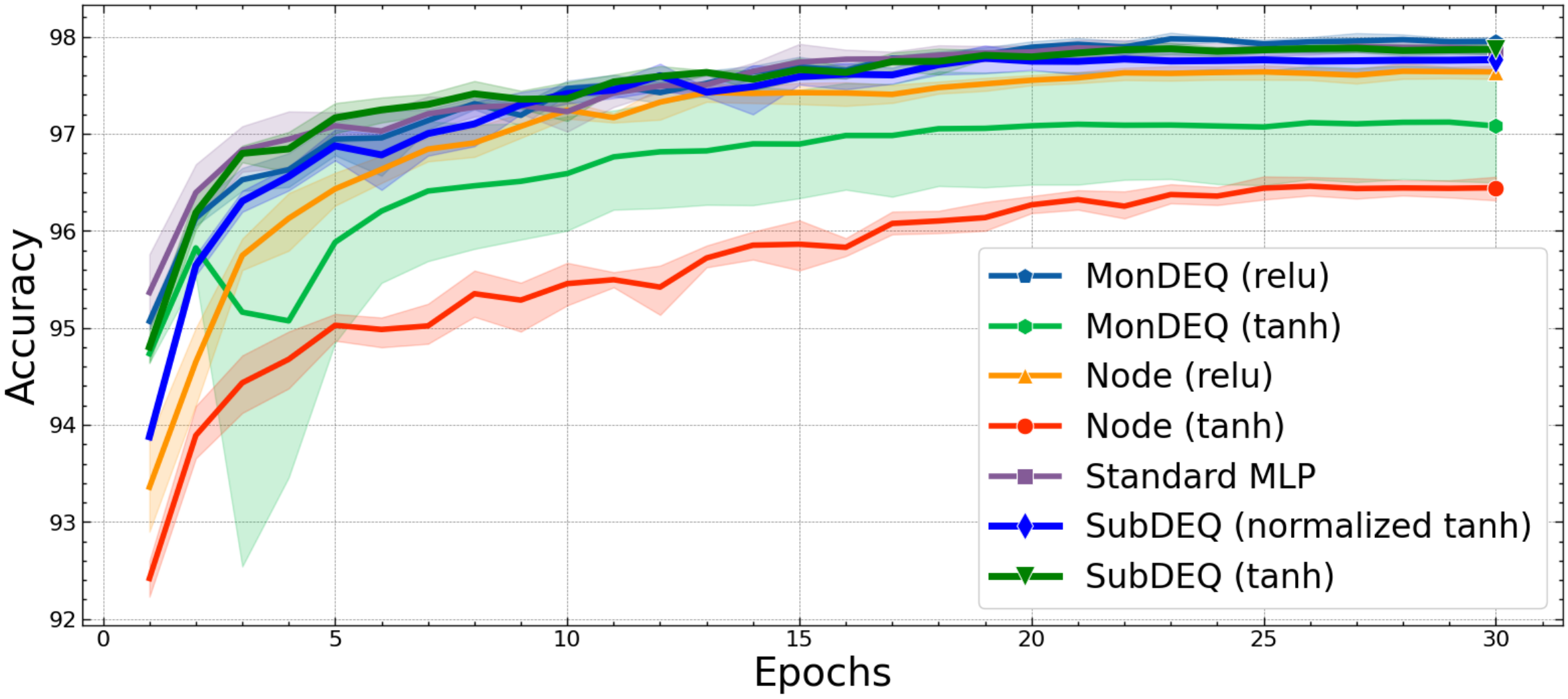}
    \caption{Validation accuracy of the dense architectures during training on MNIST.}
    \label{fig:fc_mnist}
\end{figure}

For every experiment, with the SubDEQ architectures, we use the fixed point method speed up with Anderson acceleration to calculate the fixed point and as the stopping criterion we use the relative residual 
$$
\|z_{k+1}-z_k\|_F/\|z_{k+1}\|_F,
$$
and we stop the method when the residual reaches the values of $10^{-3}$.
We decided to train the dense architecture only on MNIST, instead, we trained the convolutional models
on all three data sets.
Regarding the data, we use the hold-out approach, dividing the dataset into train validation and test sets.  \Cref{tab:hold_out_split} describes the proportion of the splittings. 
All training data is normalized to mean $\mu = 0$, standard deviation $\sigma = 1$, and the validation and the test are rescaled using the mean and standard deviation of the training data.
On the same data split we run the experiments 5 times and in  \Cref{tab:reuslt_deq} we show the mean $\pm$ the standard deviation %\ftnote{standard error or standard deviation?} 
of the misclassification error. 
As we can notice from \Cref{tab:reuslt_deq}, the SubDEQ (Normalized Tanh), depending on the dataset, achieves performance that exceeds the performance of the other model or reaches similar performance to the best model. We highlight that adding the vector $1.603$ to the activation function to ensure the uniqueness and convergence, does not negatively affect the performance, instead ensures higher stability and leads the model to a higher accuracy.

\subsection{Deep Equilibrium Graph Neural Network: nonlinear graph propagation}\label{sec:graph_deq}

\begin{table}[t]
\begin{tabular}{lll}
\hline
                                               \textbf{Dataset (\% labeled)} & \textbf{Accuracy}                         \\ \hline
\textbf{Cora citation}  \textbf{(}$\mathbf{5.2}$ \textbf{\%)}                             &                                                            \\ \hline
\textbf{APPNP}                                                     & $\mathbf{80.2027 \pm 1.9557}$\textbf{\%}  \\
APPNP (Normalized Tanh)                                      & $73.3905 \pm 2.3818$ \%                                   \\ 
APPNP (Tanh)                                      & $72.2369 \pm 2.3013$ \%                                   \\ 
\hline
\textbf{Cora author}           \textbf{(}$\mathbf{5.2}$\textbf{\%)}                             &                                                            \\ \hline
APPNP                                                     & $70.834 \pm 2.1591$ \%                                     \\
\textbf{APPNP (Normalized Tanh)}                                       & $\mathbf{73.3437 \pm 1.8252}$ \textbf{\%} \\ 
APPNP (Tanh)                                      & $72.5175 \pm 2.4537$ \%                                   \\
\hline
\textbf{CiteSeer}              \textbf{(}$\mathbf{4.2}$ \textbf{\%)}                             &                                                            \\ \hline
\textbf{APPNP}                                                    & $\mathbf{62.8625 \pm 1.4477}$ \textbf{\%} \\
APPNP (Normalized Tanh)                                       & $62.66078 \pm 1.8676 $ \%                                 \\ 
APPNP (Tanh)                                      & $62.1564 \pm 1.479$ \%                                   \\
\hline
\textbf{DBLP}                  \textbf{(}$\mathbf{4.0}$ \textbf{\%)}                             &                                                            \\ \hline
APPNP                                                     & $88.8095 \pm 0.2866$                                       \\
\textbf{APPNP (Normalized Tanh)}                                       & $\mathbf{89.4007 \pm 0.3619}$ \textbf{\%} \\ 
APPNP (Tanh)                                      & $86.87046 \pm 0.3851$ \%                                   \\
\hline
\textbf{PubMed}               \textbf{(}$\mathbf{0.8}$\textbf{\%)}                             &                                                            \\ \hline
APPNP                                                     & $77.45168 \pm 1.4433$ \%                                    \\
\textbf{APPNP (Normalized Tanh)}                                     & $\mathbf{78.5827 \pm 0.9741}$ \textbf{\%} \\
APPNP (Tanh)                                      & $77.103 \pm 1.251$ \%                                 \\  \hline
\end{tabular}\caption{Mean $\pm$ standard deviation accuracy on test set}\label{tab:reuslt_appnp}%\ftnote{bold solo il nome del method che fa meglio}}
\end{table}
We conclude with an experiment on a graph neural network architecture. Graph neural networks are typically relatively shallow due to oversmoothing and oversquashing phenomena \cite{nguyen2023revisiting,Giraldo_2023}. Thus, a fixed point implicit graph neural architecture may be ineffective. However, it is shown in  \cite{gasteiger2022predict} that the PageRank random walk on the graph may be used to propagate the input injected by a simple MLP and the limit point of the propagation leads to excellent, sometimes state-of-the-art, results. This approach, named there APPNP, is effective also because the PageRank diffusion process converges linearly to a unique PageRank vector. As the model uses the PageRank fixed point as the final latent embedding, it can be interpreted as a simple DEQ with linear activations. This is possibly a limitation of APPNP as nonlinear activations may allow for a better modeling power. Using \Cref{thm:fixed_point1} one can propose variations of this approach implementing nonlinear diffusion processes based on subhomogeneous activation functions and still maintain the same fundamental guarantees of uniqueness and linear convergence. We obtain this way a SubDEQ variation of the APPNP graph neural network model, which we discuss next. 

First, we review the standard APPNP architecture \cite{gasteiger2022predict}. Consider an undirected graph $G=(V,E)$ on $n=|V|$ nodes,  let $d_i$ be the degree of node $i$, and let $A$ be the graph normalized adjacency matrix, defined as $A_{ij}=d_i$ if $ij\in E$ and $A_{ij}=0$ otherwise. APPNP defines the node embedding $Z$ via the fixed point equation:
\begin{equation}\label{eq:appnp}
   \begin{cases}
       \tilde Z  =(1-\alpha) \tilde{A} \tilde Z+\alpha f_\theta(X) &\\
        Z = \mathrm{softmax}(\tilde Z) &
   \end{cases}
\end{equation}
where $\tilde{A}= A + I$ denotes the shifted adjacency matrix and $X \in \mathbb{R}^{n \times f}$ is the input node feature matrix.

We consider here the following nonlinear  

variations of~\eqref{eq:appnp}  
\begin{equation*}
    \begin{cases}
        \tilde Z = \tanh \big( (1-\alpha) \tilde{A} \tilde Z \big) +\alpha f_\theta(X)  + 1.2 & \\
        Z = \mathrm{softmax}(\tilde Z) &
    \end{cases}
\end{equation*}
and
\begin{equation*}
    \begin{cases}
        \tilde Z = \norm_{\|\cdot\|_{\infty}} \left( \tanh \big( (1-\alpha) \tilde{A} \tilde Z \big) +\alpha f_\theta(X)  + 1.2 \right) & \\
        Z = \mathrm{softmax}(\tilde Z) &
    \end{cases}
\end{equation*}
where the final $l^\infty$ normalization layer is implemented columnwise in the model above, as in \Cref{cor:fixed_point2}.

If we add to  $\tanh$ a translation vector with all entries equal to $1.2$, we obtain a subhomogeneous operator with degree $\mu = 0.99$, due to the \Cref{lemma:tanh+shift} and \Cref{lemma:subhom_composition}. Notice also that the Jacobian with respect ${Z}$ of this transformation is entrywise positive respect all ${Z}>0$ and is differentiable. Thus, both the nonlinear fixed point equations above have a unique fixed point due to \Cref{cor:fixed_point2}. 
We test APPNP and its nonlinear variations on different graph datasets: Cora citation, Cora author, CiteSeer, DBLP, and PubMed, always in a node classification semi-supervised learning setting.
We divide the dataset into training, validation, and test sets. The percentages of observation used for the training set are shown in  \Cref{tab:reuslt_appnp}, and the remaining observations are equally split between validation and test sets. 
 For both methods, we use similar hyperparameters, as $f_{\theta}(\cdot)$ we use $2$-layers MLP of width $64$ for each layer; we regularize the architectures with dropout $p=0.5$, we use Adam as optimizer with constant learning rate equal to $ 0.01$ and a weight decay equal to $0.005$. We set $\alpha=0.1$ and $K=10$. 
We repeat the splitting and the training 5 times and we report the average $\pm$ std results in \Cref{tab:reuslt_appnp}. 
%\ftnote{Table A??}
%\ftnote{We should try to have results also for appnp experiments in the main %paper. For the moment do not worry about space, I will then work out a way to %compress the text  if needed.  }

%CiteSeer \ctie{sen:aimag08}, 

\section{Conclusions}
We have presented a new analysis of the existence and uniqueness of fixed points for DEQ models, as well as convergence guarantees for the corresponding fixed point iterations. %introduced a new class of implicit-depth neural networks that are provably well defined in the sense that they are guaranteed to have a unique fixed point, which we can compute via a fast converging fixed point iteration. 
Unlike previous approaches that require constraints on the weight matrices to guarantee uniqueness, our theoretical framework allows us to use general weight matrices, provided the activation functions of the network are subhomogeneous. % and a final normalization layer is added to the architecture. 
We observe that many well-known activation functions are indeed subhomogeneous possibly up to some minor modification, showing the vast applicability of our framework. Thus, we provide several examples of new subhomogeneous deep equilibrium architectures designed for image classification and nonlinear graph propagation.

% \newpage

\subsection*{Impact statement}
This paper present work whose goal is to advance the field of implicit-depth deep learning architectures from a mathematical point of view. There are many pontential societal consequences of our work, none of which we feel must be specifically highlighted here.

\bibliography{references}
\bibliographystyle{icml2023}
%%%%%%%%%%%%%%%%%%%%%%%%%%%%%%%%%%%%%%%%%%%%%%%%%%%%%%%%%%%%%%%%%%%%%%%%%%%%%%%
\newpage
\appendix
\onecolumn

\section{Proofs}\label{proof_result}

\begin{proof}(\Cref{prop_eq_subhom_def})\\
We now prove the first part. Let $g \colon [1,+\infty) \to \mathbb{R}^n $, defined as
$$
g(\lambda) = [g_1(\lambda),\dots,g_n(\lambda)]=F(\lambda z)-\lambda^{\mu}F(z).
$$
Clearly, $g(1)=0$ and $g$ is a differentiable function since $F(z)>0$ for each $z>0$ and $F$ is differentiable.
\begin{align*}
g'(\lambda) &= F'(\lambda z) z-\mu \lambda^{\mu-1}F(z)\\
\lambda g'(\lambda) &= F'(\lambda z) \lambda z-\mu \lambda^{\mu}F(z) \leq | F'(\lambda z)| \lambda z - \mu \lambda^{\mu}F(z).
\end{align*}
Using the definition of subhomogeneous operator we get
$$
\lambda g'(\lambda) \leq \mu (F(\lambda z)-\lambda^{\mu} F(z)) = \mu g(\lambda).
$$
Then $g_j'(\lambda)\leq \frac{\mu}{\lambda}g_j(\lambda)$
for each $j=1,\dots,n$, thanks to the Grönwall's inequality, 
$$
g_j'(\lambda) \leq g_j(1) \exp\left(\int_{1}^{\lambda} \frac{\mu}{t}dt \right) = g_j(1) \lambda = 0.
$$
Therefore, $g_j'(\lambda)\leq 0$, this shows that $g_j$ is a decreasing function and $g_j(\lambda)\leq 0$, thus $g(\lambda)\leq0$ entry-wise.\\
We now prove the second part of the proposition, the necessary condition is implied by the first part so we will prove only the sufficient condition.\\
Let $g \colon [0,+\infty) \to \mathbb{R}^n$ be defined as 
$g(\lambda) = F(\lambda z) - \lambda^{\mu} F(z)$, by hypothesis, for each $\lambda\geq 1$, $g(\lambda)\leq 0$, also note that $g(1)=0$. This shows that  $g'(\lambda)\leq 0$ for each $\lambda \geq 1$.
$$
g'(\lambda) = F'(\lambda z)z- \mu \lambda^{\mu-1}F(z),
$$
we get
$$
g'(1) = F'(z)z- \mu F(z),
$$
which implies $ F'(z)z \leq \mu F(z)$.
\end{proof}

\begin{proof}(\Cref{thm:fixed_point1})\\
Let $D(z)$ be Clark's generalized Jacobian of the map $z \to \ln(F(e^z))$. For the Mean Theorem \cite{Clarke:Optimization_and_Nonsmooth_Analysis} we have \\
$$
\ln \left(F\left(e^{\ln(x)}\right)\right)-\ln \left(F \left(e^{\ln(y)}\right)\right) \in co\left(D(\Omega(x,y)\right)(\ln(x)-\ln(y)),
$$
where $\Omega(x,y)\colon = \{z\in\mathbb{R}^n \mid z = t\ln(x)+(1-t)\ln(y), \; t\in [0,1]\}$ and $co\left(D(\Omega(x,y)\right)(\ln(x)-\ln(y))$ denotes the convex hull of all points of the form $Z(ln(x)-ln(y))$ where $Z\in D(u)$ for some $u$ in $\Omega(x,y)$.\\
For the Caratheodory Theorem, we get
$$
\ln \left(F\left(e^{\ln(x)}\right)\right)-\ln \left(F \left(e^{\ln(y)}\right)\right) = \sum_{l=0}^{n^2} \beta_j \xi_l(\ln(x)-\ln(y)),
$$
where $\xi_l \in D(u)$ for a $u$ in $\Omega(x,y)$, $\beta_l\geq0$ and $\sum_l \beta_l=1$.
Let $v=\ln(x)-\ln(y)$, using the Chain Rule \cite{Clarke:Optimization_and_Nonsmooth_Analysis}, we obtain
$$
D(u)v \subseteq \left(\Diag(F(e^u))^{-1}\partial F(e^u) \Diag(e^u)\right)v,
$$
$\partial F(u)$ detonates the Clark's Generalized Jacobian of $F$ with respect $u$.
Therefore, we can write the element of $D(u)v$ as 
$$
\Diag(F(e^u))^{-1}Q\Diag(e^u)v,
$$
where $Q \in \partial F(u)$.
At this point, we can estimate $\delta(F(x),F(y))$ as follows,
\begin{align}\label{th:norm_est}
\delta(F(x),F(y)) &= \| \ln(F(x))-\ln(F(y)) \|_{\infty} =\|\sum_{j=0}^{n^2} \beta_j \xi_j(\ln(x)-\ln(y))\|_{\infty}\leq \notag \\
&\leq \sum_{l=0}^{n^2} \beta_l\|\xi_l\|_{\infty} \|(\ln(x)-\ln(y))\|_{\infty} \leq \max_{l=0,\dots,n^2} \|\xi_l\|_{\infty}\|(\ln(x)-\ln(y))\|_{\infty}.
\end{align}
Moreover, using the definition of subhomogeneous operator we obtain 
\begin{align*}
\|\xi_l\|_{\infty} &= \max_{i=1,\dots,n} \sum_{j=1}^n|\xi_l|_{ij} = \max_{i=1,\dots,n} \sum_{j=1}^n|\Diag(F(e^u))^{-1}Q\Diag(e^u)|_{ij} = \\ 
& = \max_{i=1,\dots,n} \sum_{j=1}^n \left|\frac{1}{F(e^u)_i} \, Q_{ij}\, e^u_j \right| = \max_{i=1,\dots,n}  \frac{1}{F(e^u)_i}  \sum_{j=1}^n \left|\, Q_{ij}\, e^u_j \right| \leq \max_{i=1,\dots,n}  \frac{1}{F(e^u)_i} F(e^u)_i \, \mu = \mu.
\end{align*}
\end{proof}

\begin{proof}(\Cref{theorem:lip})\\
For the Euler's Homogeneous Function Theorem $\varphi(z)= w_{z}^\top z$, thus $G(z)=\frac{F(z)}{w_{z}^\top F(z)}$.
Let $D(z)$ be Clark's generalized Jacobian of the map $z \to \ln(G(e^z))$. For the Mean Theorem \cite{Clarke:Optimization_and_Nonsmooth_Analysis} we have \\
$$
\ln(G(e^{\ln(x)}))-\ln(G(e^{\ln(y)})) \in co\left(D(\Omega(x,y)\right)(\ln(x)-\ln(y)),
$$
where $\Omega(x,y)\colon = \{z\in\mathbb{R}^n \mid z = t\ln(x)+(1-t)\ln(y), \; t\in [0,1]\}$ and $co\left(D(\Omega(x,y)\right)(\ln(x)-\ln(y))$ denote the convex hull of all points of the form $Z(ln(x)-ln(y))$ where $Z\in D(u)$ for some $u$ in $\Omega(x,y)$.\\
For the Caratheodory Theorem, we get
$$
\ln(G(e^{\ln(x)}))-\ln(G(e^{\ln(y)})) = \sum_{j=0}^{nm} \beta_j \xi_j(\ln(x)-\ln(y)).
$$
where $\xi_j \in D(u)$ for a $u$ in $\Omega(x,y)$, $\beta_j\geq0$ and $\sum_i \beta_i=1$.
For simplicity we will pose $v=\ln(x)-\ln(y)$. Using the Chain Rule \cite{Clarke:Optimization_and_Nonsmooth_Analysis} we obtain
$$
D(u)v \subseteq \left(\Diag(G(e^u))^{-1}\partial G(e^u) \Diag(e^u)\right)v,
$$
$\partial G(u)$ detonates the Clark's Generalized Jacobian of $G$ with respect $u$.
Since $G(x) = \frac{F(x)}{w_u^\top F(x)}$ and the generalized Jacobian of $w_u$ is zero, if we apply the chain rule \cite{Clarke:Optimization_and_Nonsmooth_Analysis} several times we obtain
$$
\partial G(e^u)\Diag(e^u)v \subseteq \left(\frac{\partial F(e^u)}{w_u^\top F(e^u)}-\frac{F(e^u)w_u^\top\partial F(e^u)}{(w_u^\top F(e^u))^2}\right)\Diag(e^u)v,
$$
the right-hand side above denote the set of points $Q\Diag(e^u)\,v$ where $Q = \frac{H}{w_u^\top F(e^u)}-\frac{F(e^u)w_u^\top K}{(w_u^\top F(e^u))^2}$ and $H,K,\in \partial F(e^u)$.\\
Therefore we can write the element of $D(u)v$ as 
$$
\Diag(G(e^u))^{-1}Q\Diag(e^u)v.
$$
At this point, we will estimate $\delta(G(x),G(y))$ using the calculations that we have done so far.
\begin{align}\label{th:norm_est}
\delta(G(x),G(y)) &= \| \ln(G(x))-\ln(G(y)) \|_{\infty} =\|\sum_{j=0}^{mn} \beta_j \xi_j(\ln(x)-\ln(y))\|_{\infty}\leq \notag \\
&\leq \sum_{j=0}^{mn} \beta_j\|\xi_j\|_{\infty} \|(\ln(x)-\ln(y))\|_{\infty} \leq \max_{j=0,\dots,mn} \|\xi_j\|_{\infty}\|(\ln(x)-\ln(y))\|_{\infty}.
\end{align}
Now we estimate the infinity norm of a matrix of the form $\Diag(G(e^u))^{-1}Q\Diag(e^u)$. Let us begin by focusing on the first two matrices of multiplication, thus
\begin{align*}
\Diag(G(e^u))^{-1}Q & =\Diag(F(e^u))^{-1}w_{u}^\top F(e^u)Q = \\
 & = \Diag(F(e^u))^{-1}H- \frac{\one w_{u}^\top K}{w_{u}^\top F(e^u)},
\end{align*}
consequently, the entries are 
\begin{align*}
\left|\Diag(G(e^u))^{-1}Q\right|_{ij}&=
\left|\Diag(F(e^u))^{-1}H- \frac{\one\ w_{u}^\top K}{w_{u}^\top P(e^u)}\right|_{ij}=\left|\frac{H_{ij}}{F(e^u)_i}-\sum_{l=1}^m\frac{w_{u,l} K_{lj}}{\sum_r w_{u,r} F(e^u)_r}\right| =
\\
& = \left|\frac{H_{ij}}{F(e^u)_i} - \sum_{l=1}^m \frac{w_{u,l} F(e^u)_l}{\sum_r w_{u,r} F(e^u)_r} \frac{K_{lj}}{F(e^u)_l}\right| = \left|\frac{H_{ij}}{F(e^u)_i}-\sum_{l=1}^{m}\gamma_l\frac{K_{lj}}{F(e^u)_{l}}\right|,
\end{align*}
where $\gamma_i = \frac{ w_{u,i} F(e^u)_i}{\sum_r w_{u,r} F(e^u)_r}$. Notice that $\gamma_i$ are positive and $\sum_i\gamma_i=1$.\\
\begin{align}\label{th:sub_est}
\| \Diag(G(e^u))^{-1}Q\Diag(e^u)\|_{\infty} &= \max_{i=1,\dots,m} \sum_{j=1}^n|\Diag(G(e^u))^{-1}Q \Diag(e^u)|_{ij} = \notag \\ 
&=\max_{i=1,\dots,m} \sum_{j=1}^n \left| \frac{H_{ij}e^u_j}{F(e^u)_i}-\sum_{l=1}^{m}\gamma_l\frac{K_{lj}e^u_j}{F(e^u)_{i}}\right| \leq \notag \\
&\leq \max_{i=1,\dots,m} \left|\frac{(H e^u)_i}{F(e^u)_i}\right|+\max_{i=1,\dots,m} \left|\frac{(K e^u)_i}{F(e^u)_i}\right| \leq \mu+\mu = 2\mu.
\end{align}
Finally we obtain $\max_{j=0,\dots,mn} \|\xi_j\|_{\infty}\leq 2 \mu$, that conclude the proof of the first part. We now assume that $F$ is differentiable and its Jacobian is entrywise positive for each $z>0$.
Until \Cref{th:norm_est} the proof is the same as the first part. Therefore, we can start estimating $\Diag(G(e^u))^{-1}Q\Diag(e^u)$ as following
\begin{align*}
    \left|\Diag(G(e^u))^{-1}Q\right|_{ij}&=
    \left|\Diag(F(e^u))^{-1}J_F(e^u)- \frac{\one w_u^\top J_F(e^u)}{w_u^\top F(e^u)}\right|_{ij}=\left|\frac{J_F(e^u)_{ij}}{F(e^u)_i}-\sum_{l=1}^m\frac{w_{u,l}J_F(e^u)_{lj}}{\sum_r w_{u,r}F(e^u)_r}\right| =
    \\
    & = \left|\frac{J_F(e^u)_{ij}}{F(e^u)_i} - \sum_{l=1}^m \frac{w_{u,l} F(e^u)_l}{\sum_r w_{u,r} F(e^u)_r} \frac{J_F(e^u)_{lj}}{F(e^u)_l}\right| =: \left|C_{ij}-\sum_{l=1}^{m}\gamma_l C_{lj} \right|
\end{align*}
 where $\gamma_i = \frac{w_{u,i}F(e^u)_i}{\sum_r w_{u,r}F(e^u)_r}$ and $C_{ij} = \frac{J_F(e^u)_{ij}}{F(e^u)_i}$. Notice that $\gamma_i$ and $C_{ij}$ are  positive and $\sum_i\gamma_i=1$.
$$
|C_{i j}-\sum_{l=1}^m \gamma_l C_{l j}| \leq \max _{l=1, \ldots,m} C_{l j}=: C_{\overline{l} j}.
$$
Thus
\begin{align*}
    \| \Diag(G(e^u))^{-1}Q\Diag(e^u)\|_{\infty} &= \max_{i=1,\dots,m} \sum_{j=1}^n|\Diag(G(e^u))^{-1}Q \Diag(e^u)|_{ij} = \\ 
    &=\max_{i=1,\dots,m} \sum_{j=1}^n|\Diag(G(e^u))^{-1}Q|_{ij} e^u_{j}\leq\\
    &\leq\sum_{j=1}^n C_{\overline{l} j}e^u_j = \sum_{j=1}^n \frac{J_F(e^u)_{\overline{l}j}}{F(e^u)_{\overline{l}}} e^u_j \leq \mu.
\end{align*}
That conclude the proof.
\end{proof}
\begin{proof}(\Cref{cor:fixed_point2})\\
Let 
$$
\delta(G(x),G(y)) = \max_{\substack{i=1,\dots,n\\
j=1,\dots,d}} |\ln(x_{ij})-\ln(y_{ij})|
$$
be the Thompson distance on $\mathbb{R}^{n \times d}$. Let $G_j:\mathbb R^{n\times d} \to \mathbb R^n$ be the $j$-th column mapping of $G$.
$$
\delta(G(x),G(y)) = \max_{\substack{i=1,\dots,n\\
j=1,\dots,d}} |\ln(x_{ij})-\ln(y_{ij})| = \max_{j=1,\dots,d}  \left(\max_{i=1,\dots,n}|\ln(x_{ij})-\ln(y_{ij})|\right) = \max_{j=1,\dots,d} \delta(G_j(x),G_j(y))\, .
$$
% where the $\delta_j$ is the thompson distance defined on $\mathbb{R}^n$. 
Applying \Cref{theorem:lip} with input space $\mathbb R^{n\times d}$ and output space $\mathbb R^n$, we obtain
$$
\delta\left(G_j(x),G_j(y)\right) \leq 2 \mu \delta(x,y)
$$
for all $j=1,\dots,d$.  Thus 
$$
\max_{j=1,\dots,d} \delta \left(G_j(x),G_j(y)\right) \leq 2 \mu \delta(x,y).
$$
This concludes the proof of the first part. Applying the same argument one can prove the case of positive Jacobian yielding the same upper bound without $2$.%reasoning we can prove the second part of the corollary.
\end{proof}

\begin{proof}(\Cref{lemma:subhom_composition}) \\
Now we start proving the first part, hence that $P \circ H$ is $ h \mu$-subhomogeneous.
Using the chain rule we obtain $ \partial (P\circ H)(z)z=\partial P(H(z)) J_{H}(z)z$, where  $\partial (P\circ H )(z)$ and $\partial P(H(z))$ are the Clarke’s generalized Jacobians of $P$ and $P\circ H$, respectively evaluated in  $z$ and $P(z)$, and $J_{H}(z)$ is the Jacobian of $H$ in $z$.\\
Therefore we can write an element of $\partial (P\circ H)(z)z$ as $MJ_{H}(z)z$, with $M \in \partial P(z)$. 
Moreover, applying Euler’s Homogeneous Function Theorem and the definition of subhomogeneity we get
\begin{align*}
        |MJ_{H}(z)z| = h |MH(z)|\leq h \mu P(H(z))= h \mu (P\circ H)(z).
\end{align*}
Thus, $P\circ H \in \subhom_{h \mu}$.\\
We now prove the second part of the Lemma, hence $Q \circ T_y \circ P \circ H$ is $ h \mu \lambda$-subhomogeneous. Let $F = Q \circ T_y \circ P \circ H$ and $\boldsymbol P = P \circ H$, for the first point is $h \mu $-subhomogeneous.
let $\partial F(z)z \subseteq co\{\partial Q( \boldsymbol P (z)+y)\partial \boldsymbol P (z)\}z$. Thus, an element $M \in\partial F(z)$ can be written as 
$$
M\,z= \sum_{k=0}^n \beta_k M^{Q}_kM^{\boldsymbol P}_k\,z,
$$
where $\beta_k \geq 0 $ and $\sum_k \beta_k = 1$, $M^{Q}_k \in \partial Q(\boldsymbol P(z)+y)$ and $M^{P}_k \in \partial \boldsymbol P(z)$.
We get
$$
|M\,z|= |\sum_{k=0}^n \beta_k M^{Q}_kM^{\boldsymbol P}_k\,z| \leq \lambda \sum_{k=0}^n \beta_k |M^{Q}_k| \boldsymbol P(z) \leq
\lambda \sum_{k=0}^n \beta_k |M^{Q}_k|(\boldsymbol P(z)+y) \leq \lambda\mu Q(\boldsymbol P(z)+y) = \lambda \mu F(z).
$$
Therefore, $Q \circ T_y \circ P \circ H \in \subhom_{h\mu\lambda}$.
\end{proof}

\begin{proof}(\Cref{lemma:sigmoid})\\
If $z\geq0$, $\sigma(z)>0$, then $\mathbb{R}_+ \subseteq \dom_{+}(\sigma)$. Also
note that 
 $$
 \sigma'(z) = \frac{e^z(1+e^z)-e^z e^z}{(1+e^z)^2} = \sigma(z)(1-\sigma(z)).
$$
Since $\sigma(z) = \frac{e^z}{1+e^z}$, the inequality that we want to prove is equivalent to $z \leq 1+e^z$.
In $0$ the inequality is verified, moreover, if we calculate the derivative of both sides we obtain
$$
1\leq e^z,
$$
$e^0 = 1$ and $e^z$ is a monotone increasing function, thus the inequality is verified. Therefore $\sigma \in \subhom_1(\mathbb{R}_{+})$.
\end{proof}

\begin{proof}(\Cref{lemma:softplus})\\
If $z \geq 0$, $\sigma(z)>0$, thus $\mathbb{R}_+ \subseteq \dom_{+}(\sigma)$.
$$
\sigma'(z) = \frac{e^{\beta z}}{1+e^{\beta z}},
$$
$0<\sigma'(z)<1$, thus $\sigma$ is a monotonic increasing function. Therefore what we want to show
coincides with 
$$
\sigma'(z)x\leq \sigma(z).
$$
Moreover, $\sigma'(z)z\leq x$, since $0<\sigma'(z)<1$. Clearly is enough to proof $\beta z \leq ln(1+e^{\beta z})$, this is equivalent to $e^{\beta z} \leq 1+e^{\beta z}$ that holds for all $z \in \mathbb{R}_+$.
\end{proof}

\begin{proof}(\Cref{lemma:tanh})\\
Note that, if $z \geq 0$, $\sigma(z)>0$, then $ \mathbb{R}_{+} \subseteq \dom_{+}(\sigma)$.
\begin{equation}\label{der:tanh}
    \sigma '(z) = \frac{4}{(e^z+e^{-z})^2}.
\end{equation}
Since, if $z \geq 0$ $\sigma '(z)>0$, what we want to prove is equivalent to $\sigma 
 '(z)\,z\leq \sigma(z)$. If we plug in the last equation the exponential formulation of the hyperbolic tangent and \ref{der:tanh} we get 
$$
\frac{4z}{e^z+e^{-z}} \leq e^z-e^{-z}.
$$
Using the exponential formulation of the hyperbolic sine we obtain
$$
2z \leq \sinh(2z).
$$
The Taylor expansion of the $\sinh$ is 
$$
\sinh(2z) =2z+\frac{(2z)^3}{3 !}+\frac{(2z)^5}{5 !}+\frac{(2z)^7}{7 !}+\cdots = \sum_{k=0}^{\infty} \frac{(2z)^{2 k+1}}{(2 k+1) !}.
$$
Since the first term of the series is $2z$ and $z \in \mathbb{R}_{+}$, the inequality is verified. Thus $\tanh \in \subhom_1(\mathbb{R}_{+})$.
\end{proof}

\begin{proof}(\Cref{lemma:hardtanh})\\
First, note that, for each $z\in \mathbb{R}$, $\sigma(z)>0$, thus $\dom_{+}(\sigma)=\mathbb{R}$.
The Clarke's generalized Jacobian of $H$ respect to $z$ is:
$$
\partial \sigma(z) = 
\begin{cases}
0  \quad \text{if} \quad z<\minval \quad \text{or} \quad z>\maxval,\\
[0,1]  \quad \text{if} \quad z=\minval \quad \text{or} \quad z=\maxval \\
1  \quad \text{otherwise}
\end{cases}
$$
Then we get 
$$
\partial \sigma(z)z = 
\begin{cases}
0  \quad \text{if} \quad z<\minval \quad \text{or} \quad z>\maxval,\\
[0,\minval]  \quad \text{if} \quad z=\minval\\
[0,\maxval] \quad \text{if} \quad z=\maxval \\
z  \quad \text{otherwise}
\end{cases}
$$
Thus by the definition of $\sigma$ we obtain the inequality $|M\,z| \leq \sigma(z)$ for each $z \in \mathbb{R}$ and for each $M \in \partial \sigma(z)$.
\end{proof}

\begin{proof}(\Cref{prop:softmax}) \\
We compute the gradient of $\sigma$ and we get 
$$
\nabla \sigma(z) = \frac{1}{\sum_i e^{z_i}}[e^{z_1},\dots,e^{z_n}]^{\top} = \frac{e^z}{\sum_i e^{z_i}}.
$$
Thanks to the convexity of the exponential function we obtain
$$
\max_i z_i = \log(e^{\max_i z_i}) \leq  \log(e^{\sum_i z_i}) \leq \log(\sum_i e^{z_i}) = \sigma(z)
$$
from the previous inequality we obtain
$$
|\nabla \sigma(z)^{\top}z| = \nabla \sigma(z)^{\top}z = \frac{\sum_i z_ie^{z_i}}{\sum_i e^{z_i}} \leq \max_i z_i \leq \sigma(z). 
$$    
This shows that $\sigma \in \subhom_1(\mathbb{R}^n_+)$. Thanks to the fact that $\nabla \sigma(z)$ is entrywise positive for each $z \in \mathbb{R}^n_+$, for \Cref{remark:sub_to_strongsub}, $\sigma \in \stronglysubhom_1(\mathbb{R}^n_+)$.
\end{proof}

%\pagebreak

\section{Additional Experiment}\label{sec:add_experiment}
We initiate by considering the general equation for a SubDEQ
$$
z = \mathrm{norm}_\varphi(\,\sigma_1(\sigma_2(Wz)+f_\theta(x))\,)\, .
$$
As we illustrate in \Cref{sec:subhomdeq}, we can build several categories of SubDEQ layer by choosing either $\sigma_1$ or $\sigma_2$ as the nonlinear activation function. The two SubDEQs proposed in \Cref{sec:experiments} have $\sigma_1 = \mathrm{Id}$ and $\sigma_2(z)$ as the activation functions, if we want to swap $\sigma_1$ and $\sigma_2$, so $\sigma_1$ be the nonlinear function and $\sigma_2$  the identity, we must restrict the hidden weights of our DEQ layers to be positive, since in order to apply \Cref{lemma:subhom_composition} $\sigma_2(Wz)$ must be positive in the positive orthant. Two examples of this kind are:
$$
\mathrm{norm}_{\|\cdot\|_{\infty}}(\tanh(|W|\,z+\mathrm{ReLU},(U\,x+b))+1.2),
$$
and exploiting the power scaling trick
$$
\mathrm{norm}_{\|\cdot\|_{\infty}}(\tanh(|W|\,z+\mathrm{ReLU}(U\,x+b))^{0.99}),
$$
where $|W|$ is meant as the absolute value applied entrywise to the weight matrix $W$. Both layers are well-posed, as their unnormalized versions are subhomogeneous with a degree of $0.99$ and differentiable with an entrywise positive Jacobian. 
We also implement a third variant of SubDEQ without the normalization using the power scaling trick, with the implicit layer defined as follows
$$
\tanh(W\,z)^{0.99}+\mathrm{ReLU}(U\,x+b),
$$

We test them on the same dataset used in \Cref{sec:experiment_image_dataset} using the same training, validation, and test splitting with the same hyperparameter of the SubDEQ models in \Cref{sec:experiment_image_dataset}, in \Cref{tab:reusult_subdeq_appendix} we report the misclassification error on the test set.

We also conduct other experiments with the graph neural network architecture. In section \Cref{sec:graph_deq} we considered the following architecture
\begin{equation}\label{eq:appnp_tanh1}
    \begin{cases}
        \tilde Z = \norm_{\|\cdot\|_{\infty}} \left( \tanh \big( (1-\alpha) \tilde{A} \tilde Z \big) +\alpha f_\theta(X)  + 1.2 \right) & \\
        Z = \mathrm{softmax}(\tilde Z) &
    \end{cases}
\end{equation}
Notice that, the map
$$
\tilde Z  \mapsto  (1-\alpha) \tilde{A} \tilde Z 
$$
is 1-subhomogeneous, $\alpha f_\theta(X)$ is a positive translation vector, and $\tanh(\cdot)+1.2$ is $0.99$-storngly subhomogeneous. Thus, $\tanh \big( (1-\alpha) \tilde{A} \tilde Z +\alpha f_\theta(X) \big) + 1.2$, due to \Cref{lemma:subhom_composition}, has a subhomogeneity degree of $0.99$, since is differentiable with an entrywise positive Jacobian, the following iteration converge 
\begin{equation}\label{eq:appnp_tanh2}
    \begin{cases}
        \tilde Z = \norm_{\|\cdot\|_{\infty}} \left( \tanh \big( (1-\alpha) \tilde{A} \tilde Z +\alpha f_\theta(X) \big) + 1.2 \right) & \\
        Z = \mathrm{softmax}(\tilde Z) &
    \end{cases}
\end{equation}
We test also this architecture on the same dataset used in \Cref{sec:graph_deq} comparing it with APPNP and the subhomogenoeus version of APPNP. Moreover, we tune the $\alpha$ parameter making a grid search on the value $[0.05,0.1,0.3,0.5,0.7,0.9]$. To tune $\alpha$ we fixed the test set as half of the unknown labels, and we trained the models by varying $\alpha$. For each model, we select the optimal $\alpha$ by choosing the one that achieved the highest accuracy on the validation set, the validation set is the half remaining part of the unknown labels. Subsequently, we trained the model five times using the optimal $\alpha$, altering the training set by sampling from all observations in the dataset that were not part of the initially fixed test set. After each training session, we measured the accuracy of the test set and reported the mean and standard deviation. The results are reported in the table \ref{tab:reuslt_appnp_appendix}, where \ref{eq:appnp_tanh2} is APPPNP (Normalized Tanh) $2$ and \ref{eq:appnp_tanh1} is APPPNP (Normalized Tanh). We tested the analogous architectures for the version without the normalization layer.
\newpage

\section{Additional table}\label{app:additional-data-results}

%\begin{table*}[h]\label{tab:hyperpam_appnp}
%\\ & \hline
%\begin{end*}[h]

\begin{table}[h]\label{tab:reusult_subdeq_appendix}
\begin{tabular}{ll}
\hline
\textbf{Model}                                    & \textbf{Error \%}                        \\ \hline
\multicolumn{2}{l}{\textbf{MNIST (Dense)}}                                                                    \\ \hline
SubDEQ ($\mathrm{Normalized Tanh}\,\,(1.603)$) & $2.088 \pm 0.1405$ \%  \\
\textbf{SubDEQ} ($\mathrm{\textbf{Tanh}}$)                         & $\mathbf{1.92 \pm 0.102}$  \textbf{\%} \\
SubDEQ ($\mathrm{Normalized Tanh}\,\, (1.2)$)                                             & $2.437 \pm 0.0788$ \%                                     \\
SubDEQ ($\mathrm{Normalized with Power scale Tanh}$)                                             & $2.568 \pm  0.0495$ \%                             \\
SubDEQ ($\mathrm{Power scale Tanh}$)                                             & $1.964 \pm  0.125$ \%                             \\
\hline
\multicolumn{2}{l}{ \textbf{MNIST (Convolutional)}}                                                            \\ \hline
SubDEQ ($\mathrm{Normalized Tanh}\,\, (1.603)$)                                    & $1.354 \pm 0.98$ \%                                       \\
\textbf{SubDEQ} ($\mathrm{\textbf{Tanh}}$)                                            & $\mathbf{0.706 \pm 0.011}$ \textbf{\%} \\
SubDEQ ($\mathrm{Normalized Tanh}\,\, (1.2)$)                                             & $1.829 \pm 0.0888 $ \%                                    \\
SubDEQ ($\mathrm{Normalized with Power scale Tanh}$)                                            & $1.773 \pm 0.1948 $ \%                                                        \\ 
SubDEQ ($\mathrm{Power scale Tanh}$)                                             & $1.316 \pm  0.0459$ \%                             \\
\hline
\multicolumn{2}{l}{\textbf{CIFAR-10}}                                                                        \\ \hline
SubDEQ ($\mathrm{Normalized Tanh}\,\,(1.603)$)                                    & $28.364 \pm 0.377$ \%                                     \\
\textbf{SubDEQ} ($\mathrm{\textbf{Tanh}}$)                                         & $\mathbf{27.946 \pm 1.7564}$ \textbf{\%}                          \\
SubDEQ ($\mathrm{Normalized Tanh} \,\,(1.2)$)                                            & $35.809 \pm 0.5953 $ \%                                     \\
SubDEQ ($\mathrm{Normalized with Power scale Tanh}$)                                               & $33.864 \pm 0.7469$ \%                                     \\   SubDEQ ($\mathrm{Power scale Tanh}$)                                             & $30.871 \pm  0.2268$ \%                             \\
\hline
\multicolumn{2}{l}{\textbf{SVHN}}                                                                            \\ \hline
\textbf{SubDEQ} ($\mathrm{\textbf{Normalized Tanh}}\,\, \mathbf{(1.603)}$) & $\mathbf{9.3562 \pm 0.2122}$ \textbf{\%} \\
SubDEQ ($\mathrm{Tanh}$)                                           & $10.3987 \pm 0.41296 $ \%                         \\
SubDEQ ($\mathrm{Normalized Tanh}\,\, (1.2)$)                                           & $25.3903 \pm 3.7394 $ \%                                  \\
SubDEQ ($\mathrm{Normalized with Power scale Tanh}$)                                             & $23.4688 \pm 4.7033 $ \%                                  \\
SubDEQ ($\mathrm{Power scale Tanh}$)                                             & $19.4077 \pm  0.1346$ \%                             \\
\hline
\end{tabular}
 \caption{Mean $\pm$ standard deviation of the misclassification error on test set }
\end{table}

\begin{table}[h]\label{tab:hold_out_split}
\begin{tabular}{llll}
\textbf{Dataset}  & Train set size   & Validation set size & Test set size     \\ \hline
MNIST    & 71 \%             & 14.5 \%              & 14.5 \%             \\
CIFAR-10 & 70 \%                & 15 \%                  & 15 \%                \\
SVHN    & 50.358 \% & 23.4235 \%  & 26.2184 \% \\\hline
\end{tabular}
\caption{Hold out split proportion}
\end{table}

\begin{table}[h]\label{tab:reuslt_appnp_appendix}
\begin{tabular}{lll}
\hline
                                               \textbf{Dataset (\% labeled)} & \textbf{Accuracy}                         \\ \hline
\textbf{Cora citation}  \textbf{(}$\mathbf{7.8}$ \textbf{\%)}                             &                                                            \\ \hline
\textbf{APPNP}                            & $\mathbf{76.3835 \pm 0.7716}$\textbf{\%}  \\
APPNP (Normalized Tanh)                                      & $73.4996 \pm 0.952$ \%        \\
APPNP (Normalized Tanh) $2$                                     & $68.978 \pm 0.6919$ \%   \\ 
APPNP (Tanh)                                      & $74.621 \pm 3.2222$ \%   \\
APPNP (Tanh) $2$                                        & $75.3079 \pm 3.1284$ \%   \\
\hline
\textbf{Cora author}           \textbf{(}$\mathbf{7.8}$\textbf{\%)}                             &                                                            \\ \hline
APPNP                          & $69.2128 \pm 2.7963$ \%                                     \\
\textbf{APPNP (Normalized Tanh)}                                       & $\mathbf{69.5713\pm 1.3922}$ \textbf{\%} \\
APPNP (Normalized Tanh) $2$                                     & $68.1995 \pm 2.2836$ \%   \\
APPNP (Tanh)                                      & $68.6672 \pm 2.8515$ \%   \\
APPNP (Tanh) $2$                                       & $67.8878 \pm 2.7859$ \%   \\
\hline 
\textbf{CiteSeer}              \textbf{(}$\mathbf{7.8}$ \textbf{\%)}                             &                                                            \\ \hline
\textbf{APPNP}                                        & $\mathbf{60.9079 \pm 1.3739}$ \textbf{\%} \\
APPNP (Normalized Tanh)                                       & $59.5334 \pm 1.1373 $ \%                                 \\ 
APPNP (Normalized Tanh) $2$                                     & $59.8739 \pm 1.6284$ \%   \\
APPNP (Tanh)                                      & $61.2358 \pm 1.2669$ \%   \\
APPNP (Tanh) $2$                                       & $60.4035 \pm 1.4294$ \%   \\
\hline 
\textbf{DBLP}                  \textbf{(}$\mathbf{6.0}$ \textbf{\%)}                             &                                                            \\ \hline
APPNP                                                     & $89.1939 \pm 0.2812$ \%                                       \\
APPNP (Normalized Tanh)                                       & $89.545 \pm 0.0526$ \% \\
APPNP (Normalized Tanh) $2$                                     & $86.9683 \pm 0.1177$ \%   \\
APPNP (Tanh)                                      & $89.4138 \pm 0.269$ \%   \\
\textbf{APPNP (Tanh)} $\mathbf {2}$                                       & $\mathbf{89.6610 \pm 0.212}$ \textbf{\%}   \\
\hline
\textbf{PubMed}               \textbf{(}$\mathbf{1.2}$\textbf{\%)}                             &                                                            \\ \hline
APPNP                                                     & $75.9526 \pm 0.76875$ \%                                    \\
APPNP (Normalized Tanh)                                     & $76.9303  \pm 0.777$ \%   \\
APPNP (Normalized Tanh)  $2$                                    & $75.0607 \pm 0.70495$ \%   \\
\textbf{APPNP (Tanh)}                                      & $\mathbf{77.9957 \pm 1.0701}$  \textbf{\%}   \\
APPNP (Tanh) $2$                                       & $77.2589 \pm 0.8646$ \%   \\
\hline
\end{tabular}\caption{Mean $\pm$ standard deviation accuracy on test set }%\ftnote{bold solo il nome del method che fa meglio}}
\end{table}

\begin{table*}[t]\label{tab:Hyperparameter}
\begin{tabular}{lllll}
\multicolumn{5}{c}{\textbf{Models Hyperparameter}}                                                             \\
                          & MNIST (Dense)    & MNIST (Conv)     & Cifar-10         & SVHN             \\ \hline
Number of input channels ($x$)      & -                & $1$              & $3$              & $3$              \\
Number of hidden channels ($z$)     & -                & $16$             & $48$             & $48$             \\
Size of hidden channels ($z$)        & -                & $28 \times 28$   & $32 \times 32$   & $32 \times 32$   \\
Hidden kernel size hidden ($z$)           & -                & $3 \times 3$     & $3 \times 3$     & $3 \times 3$     \\
Input kernel size ($x$)             & -                & $3 \times 3$     & $3 \times 3$     & $3 \times 3$     \\
Dimension of input weight matrix ($x$) & $784 \times 87$  & -                & -                & -                \\
Dimension of hidden weight matrix ($z$) & $87 \times 87$   & -                & -                & -                \\
Average pooling           & -                & $4 \times 4$     & $8 \times 8$     & $8 \times 8$     \\
Epochs                    & $30$             & $40$             & $40$             & $40$             \\
Initial learning rate     & $10^{-3}$        & $10^{-3}$        & $10^{-3}$        & $10^{-3}$        \\
Learning rate schedule    & Cosine Annealing & Cosine Annealing & Cosine Annealing & Cosine Annealing \\
minimum learning rate     & $10^{-6}$        & $10^{-5}$        & $10^{-3}$        & $10^{-3}$        \\
Weight decay              & $10^{-5}$        & $10^{-5}$        & $10^{-5}$        & $10^{-5}$        \\
Batch size                & 256              & 256              & 128              & 128              \\ \hline
\end{tabular}
\caption{Models Hyperparameter}
    \label{tab:model_hyper}
\end{table*}

\end{document}